\pdfoutput=1
\documentclass{article}

 \usepackage[preprint,nonatbib]{meta/neurips_2025}

\usepackage[utf8]{inputenc} % allow utf-8 input
\usepackage[T1]{fontenc}    % use 8-bit T1 fonts
\usepackage{hyperref}       % hyperlinks
\usepackage{url}            % simple URL typesetting
\usepackage{booktabs}       % professional-quality tables
\usepackage{amsfonts}       % blackboard math symbols
\usepackage{nicefrac}       % compact symbols for 1/2, etc.
\usepackage{microtype}      % microtypography
\usepackage{xcolor}         % colors

\usepackage{graphicx}
\usepackage{multirow}
\usepackage{amsmath}
\usepackage[capitalize,noabbrev]{cleveref}
\usepackage{arydshln}
\usepackage{tcolorbox}
\usepackage{cancel}
\usepackage[normalem]{ulem} 
\definecolor{lightblue}{RGB}{220,230,245}
\definecolor{darkblue}{RGB}{0,51,102}
\definecolor{deepgreen}{RGB}{0,128,0} 
\newtcolorbox{promptbox}{
    colback=lightblue,
    colframe=darkblue,
    boxrule=1pt,
    arc=3pt,
    fontupper=\ttfamily,
    width=\textwidth,
    left=6pt,
    right=6pt,
    top=4pt,
    bottom=4pt}
\title{Computational Reasoning of Large Language Models}

\author{
    Haitao Wu$^{1,2}$\thanks{Work is done during internship at Shanghai AI Laboratory.}
    \quad Zongbo Han$^{1}$\quad Joey Tianyi Zhou$^{3}$ \quad Huaxi Huang$^{2}$\thanks{Corresponding authors.} \quad
    Changqing Zhang$^{1}$\footnotemark[2] \vspace{4pt}\\
    $^{1}$Tianjin University \quad
    $^{2}$Shanghai AI Laboratory\\ %\vspace{8pt}
    $^{3}$A$^*$STAR Centre for Frontier AI Research\\
}

\begin{document}

\maketitle
\begin{abstract}
With the rapid development and widespread application of Large Language Models (LLMs), multidimensional evaluation has become increasingly critical. However, current evaluations are often domain-specific and overly complex, limiting their effectiveness as cross-domain proxies for core capabilities. To address these limitations and enable a unified and simple evaluation framework, an ideal proxy task should target a basic capability that generalizes across tasks and is independent of domain-specific knowledge. Turing machine provides a powerful theoretical lens by reducing complex processes to basic, domain-agnostic computational operations. This perspective offers a principled framework for evaluating foundational computational abilities essential to a wide range of tasks, particularly those involving complex, multi-step reasoning such as mathematics.
Motivated by this abstraction, we introduce \textbf{Turing Machine Bench}, a benchmark designed to assess the ability of LLMs to \textbf{strictly follow rules} and \textbf{accurately manage internal states} for multi-step, referred to as \textbf{computational reasoning}.
TMBench incorporates four key features: self-contained and knowledge-agnostic reasoning, a minimalistic multi-step structure, controllable difficulty, and a solid theoretical foundation based on Turing machine.
Empirical results demonstrate that TMBench serves as an effective proxy for evaluating computational reasoning on representative LLMs. It produces clear step-wise accuracy curves, revealing LLMs' ability to execute multi-step reasoning processes. 
By analyzing performance trends across TMBench and established reasoning benchmarks, we find strong correlations with real-world tasks, bridging real-task evaluation with basic ability assessment. 
These findings suggest that TMBench holds potential as a cross-domain dimension for evaluating reasoning in LLMs.
Code and data are available at \href{https://github.com/HaitaoWuTJU/Turing-Machine-Bench}{Repo}.
\end{abstract}

% \href{https://anonymous.4open.science/r/Turing-Machine-Bench-3B6E/README.md}{anonymous repo}
% Experimentally, we observe a strong correlation between model performance on TMBench and established reasoning benchmarks, with a Pearson correlation coefficient of 0.73. This finding supports the proposed benchmark as a unified dimension across complex LLM tasks.

\section{Introduction}
\label{sec:introduction}
Recent progress in pre-training, post-training, and scaling has significantly advanced the capabilities of LLMs. These models now demonstrate remarkable capabilities not only in traditional natural language processing tasks, such as text classification and machine translation~\cite{devlin2019bert}, but also in more complex cognitive domains, including advanced reasoning~\cite{rein2024gpqa, huang2024olympicarena}, code generation~\cite{jimenez2023swe, chen2021codex}, instruction following~\cite{la2024code}, multimodal understanding~\cite{wu2023next}, and even scientific discovery~\cite{boiko2023autonomous}. While these advancements significantly broaden the scope of potential applications, the growing complexity and diversity of cognitive tasks undertaken by LLMs present substantial challenges in accurately and reliably evaluating their true capabilities.
\textcolor{black}{The intelligence of LLMs can be mainly evaluated along two key dimensions: (1) knowledge and comprehension, and (2) reasoning and decision-making. First, knowledge and comprehension benchmarks primarily focus on the  ability of LLM to apply its internal knowledge to understand and solve user questions. These skills are typically evaluated through large-scale question answering tasks (e.g., MMLU~\cite{hendrycks2020measuring}, TriviaQA~\cite{joshi2017triviaqa}).
% These skills are typically evaluated through large-scale tasks such as question answering, textual entailment, and summarization (e.g., MMLU \cite{hendrycks2020measuring}, TriviaQA \cite{joshi2017triviaqa}).
Second, reasoning is the process of inferring and justifying conclusions from a set of premises~\cite{arkoudas2023gpt}. In particular, reasoning encompasses a broad range of tasks, including logical deduction (e.g., Big-Bench Hard~\cite{srivastava2023beyond}), mathematical reasoning (e.g., AIME~\cite{AIME2024}), formal proof (e.g., miniF2F~\cite{zheng2021minif2f}, ProofNet~\cite{azerbayev2023proofnet}), programming (e.g., HumanEval\cite{chen2021codex}) and computational reasoning(e.g., Arithmetic~\cite{lai2024executing}, CodeSimulation~\cite{la2024code}).
Specifically, computational reasoning refers to the ability to accurately interpret formal rules and execute multi-step computational operations without external tools. This ability is fundamental to modern sciences, which are generally founded on rule-based systems. Disciplines such as mathematics, physics, and chemistry rely on axioms, theorems, and laws, collectively referred to as rules. At the most fundamental level, arithmetic represents the simplest form of such rules. Although LLMs are not inherently proficient at direct numerical computation, they have demonstrated the ability to perform accurate arithmetic by simulating a Turing machine designed for arithmetic~\cite{lai2024executing}. This suggests that LLMs have the potential to faithfully execute formal rules in multi-step reasoning tasks.
By assessing the results across these two dimensions, we obtain a more comprehensive understanding of an LLM's intelligence. However, these dimensions often overlap in practice. For instance, solving a mathematical problem may still rely on background knowledge encoded during pre-training, blurring the distinction between memorization (knowledge) and computation (reasoning).}

\textcolor{black}{To systematically evaluate the computational reasoning ability of LLMs, we turn to the foundational paradigm of theoretical computer science: Turing machine. Its minimalist design and computational universality offer a precise and measurable framework to evaluate a model’s ability to follow rules and deterministic state transitions. 
We introduce TMBench, a novel benchmark for evaluating computational reasoning of LLMs by simulating the operation of an $m$-Tag Turing machine. The $m$-Tag Turing machine consists of production rules, a dynamic queue, and at each step of the process, it reads the queue head and appends production symbols corresponding to the head symbol at the queue tail, then deleting the first $m$ characters.
% provides the production rules of an $m$-Tag Turing machine along with an initial queue
% The model must identify the matching rule for the symbol under the head, then execute one precise step: rewrite the symbol, move the head left or right, and update the machine state. 
By iterating this process across multiple steps and comparing the LLM's execution trace with ground truth, we obtain a quantitative measure of its computational reasoning ability as an executor.
TMBench presents several four features: (1) Self-contained and knowledge-agnostic reasoning, as all tasks are solvable from first principles without reliance on external knowledge; (2) Minimalistic multi-step structure, deliberately designed to avoid redundancy, ensuring that each reasoning step is necessary, interpretable, and yields a verifiable intermediate result; (3) Controllable difficulty, enabled by modifying the rule sets, $m$ values, step numbers and input lengths to produce a continuum of task complexity; (4) Computational generality, as any computable function can be represented by a Turing machine, making it a principled proxy for basic abilities such as rule-following and state management.
Our main contributions are summarized as follows:
\begin{enumerate}
    \item We propose the TMBench to evaluate the computation reasoning ability of LLMs. Evaluations are conducted on a broad range of recent open-source models, from 0.6B to 671B parameters, as well as proprietary models such as Gemini, GPT, Grok, and Claude.
    \item Empirical results demonstrate that TMBench strongly correlates with established reasoning benchmarks and real-world task performance, highlighting its practical relevance. It effectively captures LLMs’ multi-step reasoning abilities through clear step-wise accuracy trends, making it a promising cross-domain metric for evaluating computational reasoning.
    \item To investigate TMBench’s characteristics, we conduct ablation studies on unbounded-step execution, decoding temperature, alphabet type, and task difficulty. Results show that LLM inevitably fails with increasing steps due to its autoregressive nature. Its stable performance across different alphabets suggests reliance on reasoning rather than statistics. Varying deletion counts enables fine-grained control of task difficulty, demonstrating the benchmark’s scalability.
\end{enumerate}}

\begin{figure}[t]
  \centering
    \includegraphics[width=1.0\linewidth]{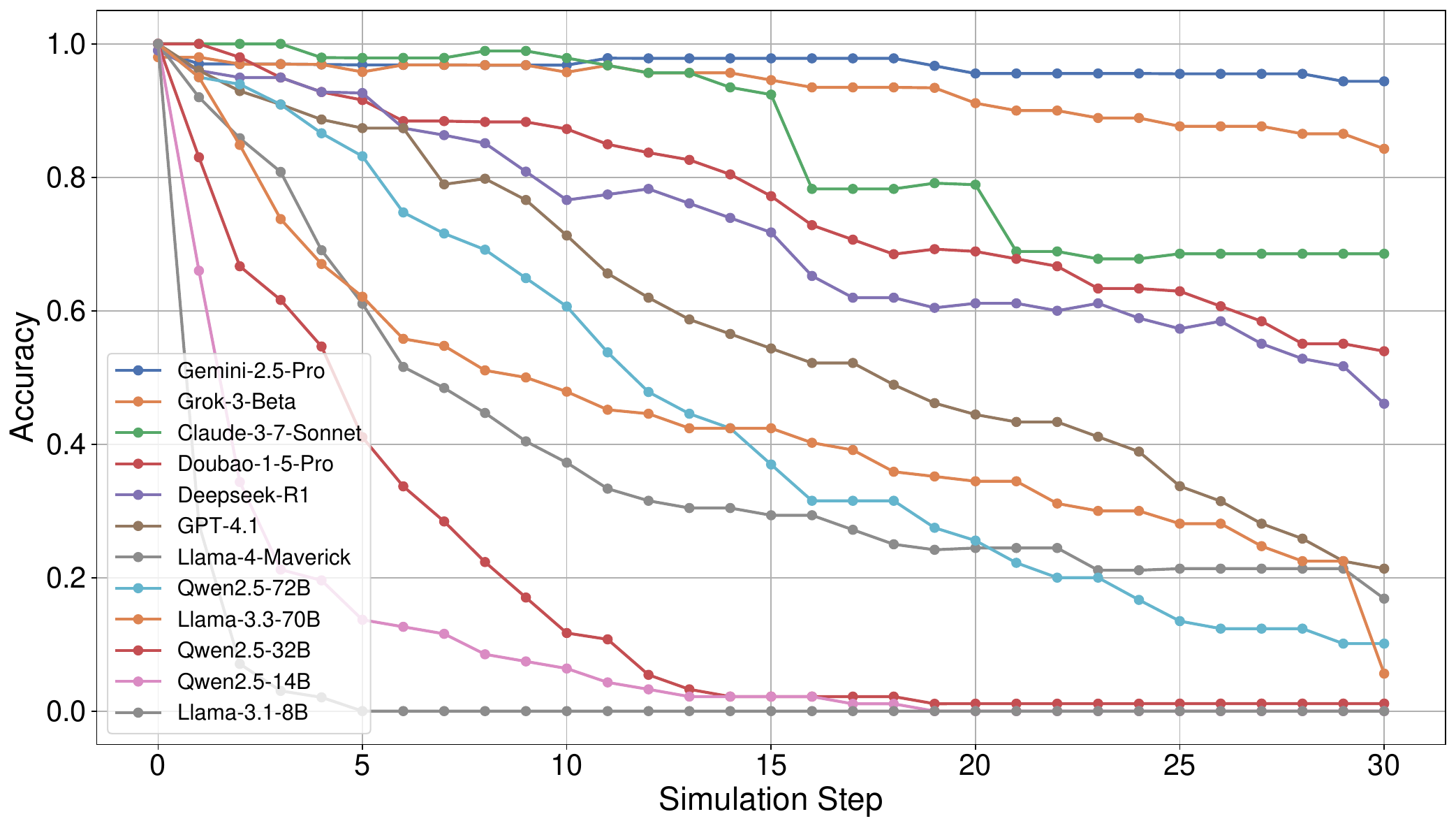}
    \vspace{-1em}
    \caption{Illustration of the multi-step performance curve on TMBench across a diverse range of both open-source and proprietary LLMs. Proprietary models demonstrate advantages in computational reasoning abilities, but accuracy still decreases as steps increase.}
    \label{fig:step_evaluation_acc}
\end{figure}
\section{Related Work}
% \subsection{Instruction Following Benchmarks}

\subsection{LLM Benchmarks}
Benchmarks for large language models can be primarily categorized into two types: closed-ended benchmarks, which provide definitive answers, and open-ended benchmarks, which rely on human preference.
For closed-ended benchmarks, a wide range of topics is addressed, including language understanding, mathematics, coding, reasoning, hallucination, toxicity, and stereotypes. Notable benchmarks in this category include MMLU~\cite{hendrycks2020measuring}, HellaSwag~\cite{zellers2019hellaswag}, GSM-8K~\cite{cobbe2021gsm}, HELM~\cite{liang2022holistic}, BigBench~\cite{srivastava2023beyond}, AGIEval~\cite{zhong2023agieval}, HumanEval~\cite{chen2021evaluating}, and ToxicChat~\cite{toxicchat}.
Beyond closed-ended questions, benchmarks also include open-ended questions that with human preference. These questions are typically assessed through expert ratings or crowd-sourced evaluations.
The recent trend includes utilizing GPT-4 for approximating human judgment~\cite{chiang-lee-2023-large}, with notable instances being MT-Bench~\cite{zheng2023judging} and AlpacaEval~\cite{alpaca_eval}.
In addition to static benchmarks, dynamic benchmarks featuring continuously updated questions are also available. These questions may be drawn from annual examinations, weekly online contests such as Codeforces~\cite{li2022competition, huang2023competition}, or collaborative platforms like ChatbotArena~\cite{zheng2023judging}.
Some studies have explored leveraging live human interaction for reinforcement learning from human preferences~\cite{bai2022training, ouyang2022training, touvron2023llama2}. 
In this paper, we introduce Turing Machine Bench, a self-contained, multi-step benchmark for computation reasoning that avoids the influence of external knowledge.

\subsection{Multi-step Reasoning}
Recent large language models demonstrate remarkable performance in solving complex reasoning tasks~\cite{lewkowycz2022solving}.
Multi-step reasoning, essential for tackling such tasks, has gained significant attention, with numerous methods proposed to enhance the reasoning capabilities of LLM.
Chain-of-Thought (CoT)~\cite{wei2022chain} and scratchpad~\cite{nye2021show} encourages step-by-step reasoning by breaking down complex problems into manageable steps, while CoT's extension, self-consistency~\cite{wang2022self}, improves performance by sampling multiple reasoning paths and selecting the most consistent solution through majority voting.
% Other approaches~\cite{21, 59, 58, 44, 48, 60, 25, 54} have further advanced this paradigm, demonstrating its effectiveness in diverse tasks.
These methods have been extensively evaluated across a variety of benchmarks, including commonsense reasoning~\cite{lourie2021unicorn,geva2021did}, mathematical reasoning~\cite{cobbe2021training,hendrycks2021measuring}, symbolic reasoning~\cite{luo2023towards,han2022folio,patel2024multi,parmar2024towards}, logical reasoning~\cite{liu2020logiqa,tafjord2020proofwriter}, and multi-modal reasoning~\cite{zellers2019recognition,xiao2021next,lu2022learn,handa2024actionreasoningbench}. These advancements have enhanced the diversity and reliability of multi-step reasoning evaluations. However, the varying levels of difficulty across steps in most reasoning tasks pose challenges for controlled assessments, particularly in quantifying error propagation and addressing context limitations. Ensuring consistent difficulty across steps is therefore crucial for transparent and reliable evaluation of the reasoning capabilities of large language models, which remains significantly underexplored.

\subsection{Rule System}
Complexity emerges from simple rules. Modern scientific subjects such as mathematics, physics, and chemistry are generally founded on rule-based systems. These rely on axioms~\cite{suppes2012axiomatic}, theorems~\cite{russell2020principles}, and laws~\cite{feynman1963feynman} to guide reasoning and model phenomena. Arithmetic is the most basic example, providing core operations and relations that support more advanced structures. In physics, Newton’s laws of motion describe how forces affect bodies, forming the basis of classical mechanics. Cellular automata illustrate how simple local rules can give rise to complex global behavior~\cite{conway1970game}.
Rule system is also essential in program verification and automated reasoning, facilitating formal proofs through Lean theorem prover~\cite{moura2021lean}, for example.
We define the ability of an LLM to faithfully execute rules step by step as computational reasoning. Based on this definition, we propose TMBench, a rule-based system grounded in Turing machines. In TMBench, LLM is prompted to output the result of each rule execution step, which is then evaluated against ground-truth trajectories for assessment. Accurate intermediate steps are essential to the integrity of the rule-based system.

% However, multi-step reasoning is indispensable for addressing complex tasks, which is widely regarded as a fundamental component of intelligence.

\section{Turing Machine Bench}
\label{sec:tmbench}
\subsection{Motivation}
% Numerous benchmarks have been proposed to assess the reasoning abilities of LLMs. Among reasoning methods, deductive reasoning, which is central to rigorous proof, is about selecting rules and applying them faithfully.
Reasoning is one of the most important abilities of LLMs and has gain significant attention recently.
Current reasoning benchmarks focus on the final results and depend on domain-specific knowledge. For instance, the AIME benchmark, used to evaluate mathematical reasoning abilities, relies on the mastery of mathematical knowledge.
Furthermore, there is a risk that models might tailor their responses to specific benchmark metrics, thereby inflating their performance.
From an intuitive perspective, reasoning is the process of \textbf{selecting} appropriate rules and \textbf{applying} them faithfully step by step.
% , for example, the classic syllogism of Socrates.
We refer to this capability as \textit{computational reasoning}.
Computational reasoning is the ability to systematically select and accurately apply rules, ensuring that each step is transparent, verifiable, and grounded within the given rule system. In contrast to logical reasoning, which primarily focuses on the validity of conclusions, computational reasoning emphasizes the faithfulness and traceability of the entire reasoning process. Therefore, we propose an evaluation framework to evaluate computational reasoning by Universal Turing Machine Simulation.

\subsection{Turing Machine}
\textbf{Turing Machine (TM)} is a foundational model of computation~\cite{turing1936computable}, defining algorithmic processes and computational limits. Formally, a TM is represented as \( \mathcal{T} = (Q, \Gamma, \delta, q_0, H) \), where \( Q \) is the set of states, \( \Gamma \) the tape alphabet, \( \delta: Q \times \Gamma \to Q \times \Gamma \times \{L, R\} \) the transition function, \( q_0 \) the initial state, and \( H \) the set of halting states.
A TM manipulates symbols on an infinite tape using a read-write head, guided by a transition function $\delta$, and executes computations step by step until it reaches a halting state.
It serves as a fundamental abstraction for formalizing decidability and complexity classes.
Beyond theoretical significance, TM are valuable for modeling symbolic reasoning and structured inference. In this work, we leverage Turing Machine principles to construct a rigorous multi-step benchmark, evaluating LLMs' ability to perform structured computations which is vital for reasoning.

% is a theoretical model of computation~\cite{turing1936computable} formally defined as \( \mathcal{T} = (Q, \Gamma, \delta, q_0, H) \), where \( Q \) is the finite set of states, \( \Gamma \) is the tape alphabet, \( \delta \) is the transition function, \( q_0 \) is the initial state, and \( H \) represents the set of halting states. Turing machine operates with an infinite tape and a read-write head that can manipulate symbols on the tape according to a set of transition rules. 

\textbf{Universal Turing Machine (UTM)} is a theory model of computation, capable of simulating the behavior of any other Turing machine. As a foundation of computability theory, the UTM formalizes the notion of computational universality and provides the theoretical foundation for the Church-Turing thesis, which asserts that any effectively computable function can be executed by a Turing Machine. By encoding both a machine's description and its input on its tape, a UTM demonstrates that a single device can emulate any computational process, establishing the basis for general-purpose computation and modern computing architectures.

To facilitate theoretical analysis while preserving computational expressiveness, we adopt the \(m\)-Tag System, a simple and recognized model. The \(m\)-Tag System has been rigorously proven to be Turing-complete for \(m > 1\)~\cite{Wang1971-WANTSA-3,UTM-2tag}, making it a suitable abstraction for modeling universal computation within a structured and analyzable framework.
% is capable of simulating the behavior of any other Turing machine. UTM is foundational to the theory of computation, as it demonstrates the universality of computation and underpins the Church-Turing thesis. The UTM illustrates that a single machine, when provided with an appropriate program and input, can simulate the behavior of any other computational machine, thereby establishing the principle of computational universality. For simplicity, we employ the \(m\)-Tag System, which has been proven to be Turing-complete when \(m > 1\)~\cite{Wang1971-WANTSA-3,UTM-2tag}.

\textbf{m-Tag System} is a formal computational model introduced in~\cite{post1943formal} as a simplified yet computationally equivalent variant of the Universal Turing Machine. It operates on a queue of symbols, iteratively applying production rules to modify the sequence. A tag system is formally described by a triplet \( (m, A, P) \), where:
\begin{itemize}
    \item \( m \) is the deletion number, specifying how many symbols are removed from the head of queue per step.
    \item \( A \) is a finite alphabet of symbols, from which queue elements are drawn.
    \item \( P \) is a set of production rules mapping each \( x \in A \) to a corresponding word \( P(x) \), which is appended to tail of the queue.
\end{itemize}

The single-step process is formally defined as follows:
% \begin{equation}
% Step : \underset{\text{Read}}{\underset{\rule{1.5em}{0.8pt}}{x_1}} x_2 \dots x_m X \to \underset{\text{Delete}} {\underset{\rule{4em}{0.8pt}}{\text{\sout{$x_1 x_2 \dots x_m$}}}} \, X \underset{\text{Write}} {\underset{\rule{2.2em}{0.8pt}}{P(x_1)}}, 
% \end{equation}
\begin{equation}
\text{Step} : \underset{\text{Read}}{\underline{x_1}} x_2 \dots x_m\, X 
\ \longrightarrow \ 
\underset{\text{Delete}}{\textcolor{white}{\underline{\textcolor{black}{\text{\sout{$x_1 x_2 \dots x_m$}}}}}}\, X 
\, \underset{\text{Write}}{\underline{P(x_1)}},
\end{equation}
where new symbols $P(x_1)$ are appends to the tail of queue generated based on the head symbol $x_1$ and production rule.
Simultaneously, $ m $ symbols are deleted from the head of the queue.
This process resembles the next-token prediction mechanism employed by LLMs.
% It is similar to basic principle of LLM: Next Token Prediction.

Tag systems are proven Turing completeness when \( m > 1 \), making them the minimal yet effective computational models, see~\cref{sec:prove} for complete certification.
An example of a 2-tag system simulation is provided in~\cref{tab:2tag_example}, illustrating the iterative process of reading, writing, and deletion until a halting condition is reached.

% \begin{table}[h!]
% \centering
% \begin{tabular}{|l|l|l|}
% \hline
% \textbf{Alphabet} & \{A, B, C, D, E\} \\ \hline
% \textbf{Init Queue}     & [D D D D]      \\ \hline
% \textbf{P-Rules} & 
% \begin{tabular}[c]{@{}l@{}}
% A : D B A B C \\
% B : B \\
% C : B \\
% D : A \\
% E : E D B
% \end{tabular} \\ \hline
% \textbf{Steps} & 
% \begin{tabular}[c]{@{}l@{}}
% 1. [D D D D] (Init) \\
% 2. [\sout{D D} D D A]\\
% 3. [\sout{D D} A A]\\
% 4. [\sout{A A} D B A B C]\\
% \end{tabular} \\ \hline
% \end{tabular}
% \vspace{0.5em}
% \caption{Example of a 2-tag system simulation. At each step, the head symbol of the queue is read, and new symbols are appended to the tail based on production rules. The first two symbols are then removed. The system halts when the queue contains fewer than two symbols.}
% \label{tab:2tag_example}
% \end{table}

\begin{table}[h!]
\centering
\renewcommand{\arraystretch}{1.15}
\caption{Example of 2-tag systems simulation. At each step, the head symbol of the queue is read, and new symbols are appended to the tail based on production rules. The first two symbols are then removed. The system halts when the queue contains fewer than two symbols.}
\begin{tabular}{|l|l|l|l|}
\hline
\textbf{Alphabet} & \{A, B, C, D, E\}& \{1, 2, 3, 4, 5\} &\{@, \#, \$, \%, \&\}  \\ \hline
\textbf{Init} & {[B A E E C]} & {[5 2 3 2]} & {[\$ @ @ \#]} \\ \hline
\textbf{P-Rules} & 
\begin{tabular}[t]{@{}l@{}}
A : E D A B C \\
B : D \\
C : E E E D D \\
D : B C \\
E : D
\end{tabular} & 
\begin{tabular}[t]{@{}l@{}}
1 : 5 5 2 \\
2 : 4 2 5 1 3 \\
3 : 4 3 1 \\
4 : 3 4 \\
5 : 3
\end{tabular} & 
\begin{tabular}[t]{@{}l@{\,:\,}l@{}}
@ & \% \$ \# \$ \\
\# & \& \\
\$ & \& \\
\% & \# \# \% \% \\
\& & \% \# \& \$ \\
\end{tabular} \\ \hline
% \begin{tabular}[t]{@{}l@{}}
% @ : \% \$ \# \$ \\
% \# : \& \\
% \$ : \& \\
% \% : \# \# \% \% \\
% \& : \%\#\&\$
% \end{tabular} \\ \hline
\textbf{Steps} & 
\begin{tabular}[t]{@{}l@{}}
0. [B A E E C] (Init) \\
1. [\sout{B A} E E C D] \\
2. [\sout{E E} C D D] \\
3. [\sout{C D} D E E E D D] \\
... \\
16. [\sout{B C} D D] \\
17. [\sout{D D} B C] \\
18. [\sout{B C} D] (Halt) \\
\end{tabular}
& 
\begin{tabular}[t]{@{}l@{}}
0. [5 2 3 2] (Init) \\
1. [\sout{5 2} 3 2 3] \\
2. [\sout{3 2} 3 4 3 1] \\
3. [\sout{3 4} 3 1 4 3 1] \\
... \\
28. [\sout{51} 3343455243134552343] \\
29. [\sout{33} 43455243134552343431] \\
30. [\sout{43} 45524313455234343134]
\end{tabular} 
& 
\begin{tabular}[t]{@{}l@{}}
0. [\$ @ @ \#] (Init) \\
1. [\sout{\$ @} @ \# \&] \\
2. [\sout{@ \#} \& \% \$ \# \$] \\
3. [\sout{\& \%} \$\#\$\%\#\&\$] \\
... \\
28. [\sout{\# \&} \$\%\#\&\$\&]\\
29. [\sout{\$ \%} \# \& \$ \& \&] \\
30. [\sout{\# \&} \$ \& \& \&]
\end{tabular} \\ \hline
\end{tabular}
\label{tab:2tag_example}
\end{table}

\subsection{Evaluation Metrics}
To comprehensively evaluate the multi-step instruction following capability of LLMs in the reasoning process, we define three key metrics: Step Accuracy, Step-Weighted Accuracy, and Pass Rate.

\paragraph{Step Accuracy.} It quantifies the proportion of correctly predicted queues at a given step in the reasoning process, providing a fine-grained evaluation of a model’s stepwise performance. It is particularly valuable for multi-step reasoning tasks, where errors can accumulate and propagate, significantly impacting overall accuracy. 
% By assessing correctness at each step, Step Accuracy offers insights into a model’s ability to navigate intermediate uncertainties, maintain logical consistency, and ensure stable reasoning trajectories.
The accuracy at step $i$ is defined as:

\begin{equation}
\text{ACC(i)} = \frac{N_{\text{correct}}(i)}{N_{\text{total}}(i)} ,
\end{equation}

where \(N_{\text{correct}}(i)\) and \(N_{\text{total}}(i)\) denote the number of correct predictions and the total number of predictions at step \(i\), respectively.

% \paragraph{Reasoning AUC (R-AUC).} It quantifies the cumulative accuracy of a model over the entire reasoning process by measuring the Area Under the step-accuracy Curve. 
% % It tracks the model's step-wise accuracy (\textit{ACC\textsubscript{step}}) as a function of reasoning steps (\(t\)), providing a single scalar value to represent overall performance and efficiency.
% R-AUC is formally defined as:
% \begin{equation}
% \text{R-AUC@T} = \int_{0}^{T} \text{ACC(i)}  \, dt \approx \sum_{t=1}^{T} \text{ACC(i)} \cdot \Delta t,
% \end{equation}
% where \(T\) is the accumulated reasoning steps, and $\Delta t = \frac{1}{T}$ for equal step intervals.
% Reasoning AUC captures the instruction following accuracy of LLMs' reasoning trajectory. A higher R-AUC indicates sustained accuracy throughout the reasoning process, conversely, a low R-AUC score reflects a rapid deterioration of accuracy with increasing reasoning depth. 

\paragraph{Step-Weighted Accuracy (SWA)}
To evaluate model performance across a sequence of reasoning steps with an emphasis on later steps, we define the \emph{Step-Weighted Accuracy} as a weighted average of per-step accuracies:
\begin{equation}
\text{SWA} = \frac{1}{\sum_{i=1}^{T} w_i} \sum_{i=1}^{T} w_i \cdot \text{ACC}(i),
\end{equation}
where $w_i$ is the weight assigned to step $i$. Setting $w_i = 1$ corresponds to uniform weighting across all steps, while $w_i = i$ linearly increases the emphasis on later steps, thereby prioritizing accurate performance in deeper stages of reasoning. 
We evaluate performance using Step-Weighted Accuracy (SWA) under both uniform (Uni.) and linear (Lin.) weighting schemes.

\paragraph{Pass Rate.}
The pass rate quantifies the likelihood of successfully completing a given process without errors before termination or reaching the maximum number of allowed steps. This metric focuses more on the final outcome, whereas the previous metrics emphasize the process.
\begin{table*}[t!]
\centering
\caption{Performance of large language models across different scales (0.6B to 605B) and API-only models, evaluated on TMBench. Asterisks (*) denote widely recognized reasoning benchmarks.}
\begin{tabular}{lcccccc}
\toprule
\textbf{Model} & \textbf{SWA (Uni.)} & \textbf{SWA (Lin.)} & \textbf{Pass Rate} & \textbf{AIME*} & \textbf{MATH*} & \textbf{GPQA*}\\
\midrule
\multicolumn{7}{l}{\textbf{0.6B - 8B}} \\
Qwen3-0.6B &0.9&0.1&0&6.7&65.0&21.7 \\
Llama-3.2-1B &3.3&0.3&0&0&18.4&12.6 \\
% Llama-3.2-3B &3.3&0.3&0&6.7&42.0 &24.7\\
Qwen3-1.7B &1.5&1.2&1&26.7&86.4&32.8 \\
Qwen3-4B &7.4&3.5&6&56.7&91.6&49.0 \\
Llama-3.1-8B &4.5&0.4&1&0&46.4&21.2 \\ 
Qwen3-8B &7.2&4.4&8&60.0&91.4&56.6\\[0.5pt]
\hdashline
\addlinespace[2.0pt]
\multicolumn{7}{l}{\textbf{12B+}} \\
Gemma-3-12B &7.7&1.6&0&13.3&83.2&41.9\\
Qwen2.5-14B &10.3&2.6&0&16.7&79.2&43.9\\
R1-Distill-Qwen-14B &29.5&18.0&10&60.0&87.0&53.5\\[0.5pt]
Qwen3-14B &13.8&9.3&2&76.7&93.0&64.1 \\ [0.5pt]
\hdashline  %%\\[-10pt]
\addlinespace[2.0pt]
\multicolumn{7}{l}{\textbf{27B+}} \\
Gemma-3-27B &26.1&12.2&5&40.0&88.8&46.0\\
Qwen2.5-32B &18.2&6.0&7&10.0&80.6&48.0\\
Sky-T1-32B-Preview &20.9&9.1&8&40.0&86.2&51.0\\
QwQ-32B-Preview &19.8&9.8&4&53.3&90.4&62.1\\
R1-Distill-Qwen-32B &33.5&22.5&10&72.6&94.3&62.1\\
Qwen3-32B &10.5&8.3&7&70.0&94.8&63.6 \\
Qwen3-30B-A3B &56.7&46.8&16 &80.4 &95.9 &65.8 \\[0.5pt]
\hdashline
\addlinespace[2.0pt]
\multicolumn{7}{l}{\textbf{70B+}} \\
Llama-3.3-70B &45.2&34.3&12&20.0&75.8&43.9\\
QVQ-72B-Preview &14.4&6.4&3&33.3&82.8&52.5\\
Qwen2.5-Math-72B &42.1&26.6&15&20.0&85.2&46.0\\ 
Qwen2.5-72B &45.6&29.6&19&13.3&82.8&45.5   \\
Llama-4-Scout &7.7&1.2&1&28.3&84.4&58.7 \\
Llama-4-Maverick &39.1&27.8&19&39.0&88.9&67.1 \\
DeepSeek-V3  &87.4&84.9&82&52.0&94.2&65.5\\ 
DeepSeek-R1  &72.2&63.8&45&79.8&97.3&71.5\\
Qwen3-235B-A22B  &45.2&32.6&22&85.7 &93.0 &70.0 \\
% Qwen3-235b-a22b \\

\hdashline
\addlinespace[2.0pt]
\multicolumn{7}{l}{\textbf{API only}} \\
% GPT-4o-mini &&&&-&70.2&40.2\\
Qwen-2.5-Max &27.0&11.8&7&23.3&83.5&58.7 \\
OpenAI-O1-mini &37.0&21.6&11&63.6&90.0&60.0 \\
Gemini-1.5-Pro &40.3&18.4&10&20.0&84.2&56.6 \\ 
% OpenAI-o1 &&&&79.2&96.4&75.7\\
OpenAI-O3-mini &37.1&30.5&7&87.3&98.5&79.7\\
% OpenAI-o3 &&&&&&\\
GPT-4.1 &58.7&45.7&26&48.1&91.3&66.3\\
Doubao-1.5-Pro &76.9&69.1&54&33.3&88.6&65.0 \\
Claude-3.7-Sonnet &85.1&78.1&69&61.3&96.2&78.2 \\
Grok-3-Beta &94.6&92.4&86&83.9&94.8&80.2 \\
% \textbf{Gemini-2.5-Pro} &\textbf{96.6} &\textbf{96.2} &\textbf{94} & \textbf{92.0} & \textbf{98.9} & \textbf{84.0}  \\
Gemini-2.5-Pro &96.6 &96.2 &94 & 92.0 & 98.9 & 84.0 \\
\bottomrule
\end{tabular}
% Models under 8B parameters struggle with even the simplest reasoning steps, consistent with emergent abilities~\cite{wei2022emergent}.
\label{tab:model_performance}
\end{table*}

\section{Experiments and Results}
\label{sec:exp_res}
In this section, we present a series of experiments conducted on TMBench using a diverse set of models and analytical approaches. The experiments and their results are outlined as follows. First, we introduce the experimental setup including datasets and LLMs. Next, we evaluate a broad range of recent open-source LLMs, spanning from 0.6B to 671B parameters, alongside proprietary models such as Gemini, Grok, and Claude. We then assess the correlation between the TMBench Pass Rate and real-world benchmarks. Finally, we conduct ablation studies including unbounded-step, temperature, alphabet, and difficulty ablations to analyze TMBench's characteristics, which inevitably fail with increasing steps due to its autoregressive nature, remain robust across different alphabets, and allow continuous and scalable control of task difficulty.
\subsection{Experimental Setup}
We provide a brief overview of the experimental setup, including the datasets and the latest large language models evaluated. Detailed information is provided in the released code and data.

\textbf{Datasets.} Based on the dataset design methodology introduced in~\cref{sec:tmbench}, we sample 100 instances of m-tag systems with an alphabet size of 5, where m = 2. For each system, the rule lengths range from 1 to 5, and the initial string lengths range from 2 to 9. The maximum simulation length is 30, with 11 cases halting early upon meeting the termination condition.
Experiments under other parameter settings can be found in Section~\ref{sec:ablation}. In addition, we evaluate model performance on several established benchmarks, including AIME2024~\cite{AIME2024}, MATH500~\cite{lightman2023lets}, GPQA Diamond~\cite{rein2024gpqa}, and MMLU Pro~\cite{wang2024mmlu}. Further details can be found in the Appendix.
% Details are provided in~\cref{tab:tm-detail}.
% \input{tabs/tm_detail}

\textbf{LLMs.} We evaluate a diverse set of state-of-the-art LLMs, including open-source models ranging from 0.6B to 671B parameters, as well as proprietary API-only models, covering various model families and architectures to ensure a comprehensive analysis.
Our selection includes the instruct-tuned versions of the LLaMA models, specifically the 1B, 8B, and 70B variants, released by the LLaMA team~\cite{dubey2024llama}. From the Qwen family, we evaluate both the instruct and preview versions of the 0.6B, 1.7B, 4B, 8B, 14B, 32B, and 72B models~\cite{yang2024qwen2}, along with the language model component of the QVQ-72B-Preview multimodal system~\cite{wang2024qwen2,qvq-72b-preview}, as well as two MoE models: Qwen3-30B-A3B and Qwen3-235B-A22B.
For the Gemma family, we include both Gemma3-12B and Gemma3-27B~\cite{team2025gemma}.
We also incorporate the QwQ-32B-Preview and Sky-T1-32B-Preview model~\cite{team2025sky}. For proprietary models, including Gemini-2.5-Pro\cite{gemini2025pro}, Grok-3~\cite{grok3}, Claude-3.7-Sonnet-Thinking~\cite{claude37sonnet}, Doubao-1.5-Pro~\cite{doubao15pro}, and GPT-4.1~\cite{openai-o1-systemcard}.
For efficient and high-throughput inference, we leverage the Transformers library~\cite{wolf2020transformers} alongside vLLM~\cite{kwon2023efficient}, which provides optimized execution for large-scale model evaluations. Unless otherwise specified, we employ a greedy decoding strategy and set the maximum generation length to 16384 tokens.

\subsection{Evaluation of Computational Reasoning}
% \textbf{Multi-step Reasoning Assessment. (R1).}
We evaluate the computation reasoning capabilities of the latest LLMs by analyzing their performance across three metrics: \textit{SWA Uniform}, \textit{SWA Linear}, \textit{Pass Rate}. These metrics offer a comprehensive evaluation of the models' ability to faithfully execute multi-step reasoning.
Detailed results are summarized in~\cref{tab:model_performance}, and the accuracy–step curves are shown in~\cref{fig:step_evaluation_acc}.
We observe that Gemini-2.5-Pro exhibits robust accuracy in computational reasoning, achieving over 90\% accuracy at 30 steps, which suggests an emergent ability to simulate Turing Machine.
It is found that models smaller than 4B struggle with even the first step. This observation further supports the phenomenon of \textbf{emergence}~\cite{wei2022emergent}.
The \textit{QVQ} multimodal model exhibits a notable decline, potentially attributable to its multimodal training process.

\subsection{Correlation with Established Benchmarks}

\begin{figure}[t]
  \centering
    \includegraphics[width=1.0\linewidth]{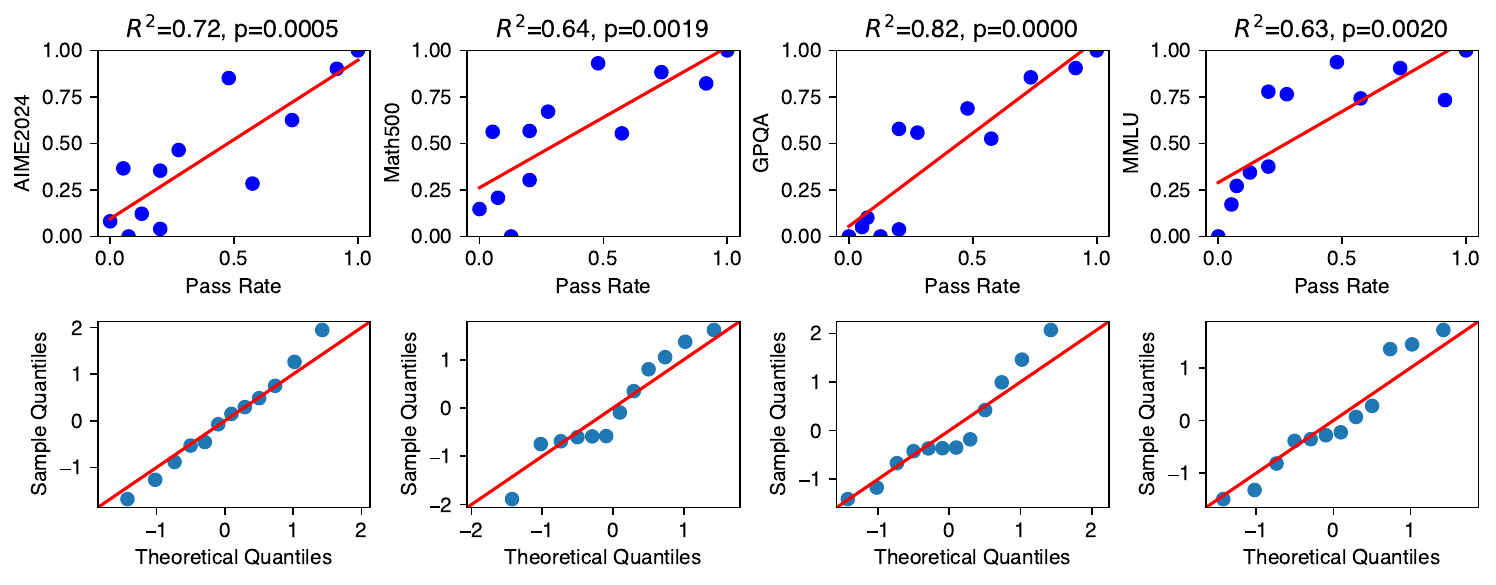}
    % \vspace{-1em}
    \caption{Correlation between TMBench Pass Rate (computational reasoning) and established benchmarks (AIME2024, MATH500, GPQA Diamond and MMLU Pro), where both metrics are min-max normalized, across 12 leading LLMs. Top: Scatter plots with linear fit between TMBench and each benchmark. Bottom: Q-Q plots of regression residuals.}
    \label{fig:corr_42}
\end{figure}
% Computation Reasoning as a basic dimension for evaluation
To investigate the relationship between computational reasoning ability and real-task reasoning performance, we analyzed the correlation between TMBench Pass Rate and established benchmarks (AIME2024, MATH500, GPQA Diamond and MMLU Pro) across 12 leading LLMs, as shown in~\cref{fig:corr_42}. The evaluated models include Gemini-2.5-Pro, Grok-3-beta, Claude-3-7-Sonnet, Doubao-1-5-Pro, DeepSeek-R1, GPT-4.1, Llama-4-Maverick, Qwen2.5-72B, Llama-3.3-70B, Qwen2.5-32B, Gemma-3-27B, and Qwen2.5-14B. The results reveal statistically significant correlations between TMBench and real-task benchmarks (p < 0.05), confirming the validity of computational reasoning as a predictor of real-world performance. Among the benchmarks, the strength of correlation follows the expected order: GPQA > AIME2024 > MATH500 > MMLU. These findings indicate that tasks with higher demands for deep reasoning and computation exhibit stronger correlations with TMBench performance. In contrast, benchmarks like MMLU, which rely more on factual knowledge than reasoning, show lower R square.
The bottom Q-Q plots of regression residuals show that the residuals are approximately normally distributed, supporting the assumption of normality in the regression model.
Moreover, we compute the average score across AIME2024, MATH500, and GPQA Diamond, obtaining a Pearson correlation coefficient of 0.882 with the TMBench pass rate, as shown in~\cref{fig:correlation}. This strong correlation suggests that advanced reasoning performance is closely associated with computational reasoning ability.

% \subsection{Concerns regarding Recognized Benchmarks}
% As illustrated in~\cref{fig:correlation}, some models, such as Llama-4-Maverick, demonstrate significantly stronger performance on AIME2024, MATH500, and GPQA compared to TMBench, raising significant concerns. TMBench is designed to evaluate basic multi-step reasoning and simple symbolic manipulation. A striking example is that certain LLMs fail at basic calculations, calling into question their ability to perform more complex mathematical reasoning. This discrepancy suggests potential risks of data leakage and overfitting.

% \begin{table}[ht!]
% \centering
% \begin{tabular}{lccc}
% \hline
% \textbf{Dataset} & \textbf{SWA (Uni.)} & \textbf{SWA (Lin.)} & \textbf{Pass Rate} \\
% \hline
% AIME       &  --  &  --  &  -- \\
% Math500    &  --  &  --  &  -- \\
% GPQA       &  --  &  --  &  -- \\
% LiveBench  &  --  &  --  &  -- \\
% BIG-Bench  &  --  &  --  &  -- \\
% MMLU       &  --  &  --  &  -- \\
% \hline
% \end{tabular}
% \caption{Evaluation metrics across datasets.}
% \label{tab:evaluation-metrics}
% \end{table}

% \paragraph{Correlation with other Benchmarks. instruction following, coding, understanding} 

% \subsection{Robustness to Data Leakage}

% BASE
% train num: 100
% train num: 200
% train num: 300
% train num: 400
% train with test

% \subsection{Controllable Difficulty}

\subsection{Ablation Study}
\label{sec:ablation}
\begin{figure}[t]
  \centering
    \includegraphics[width=0.99\linewidth]{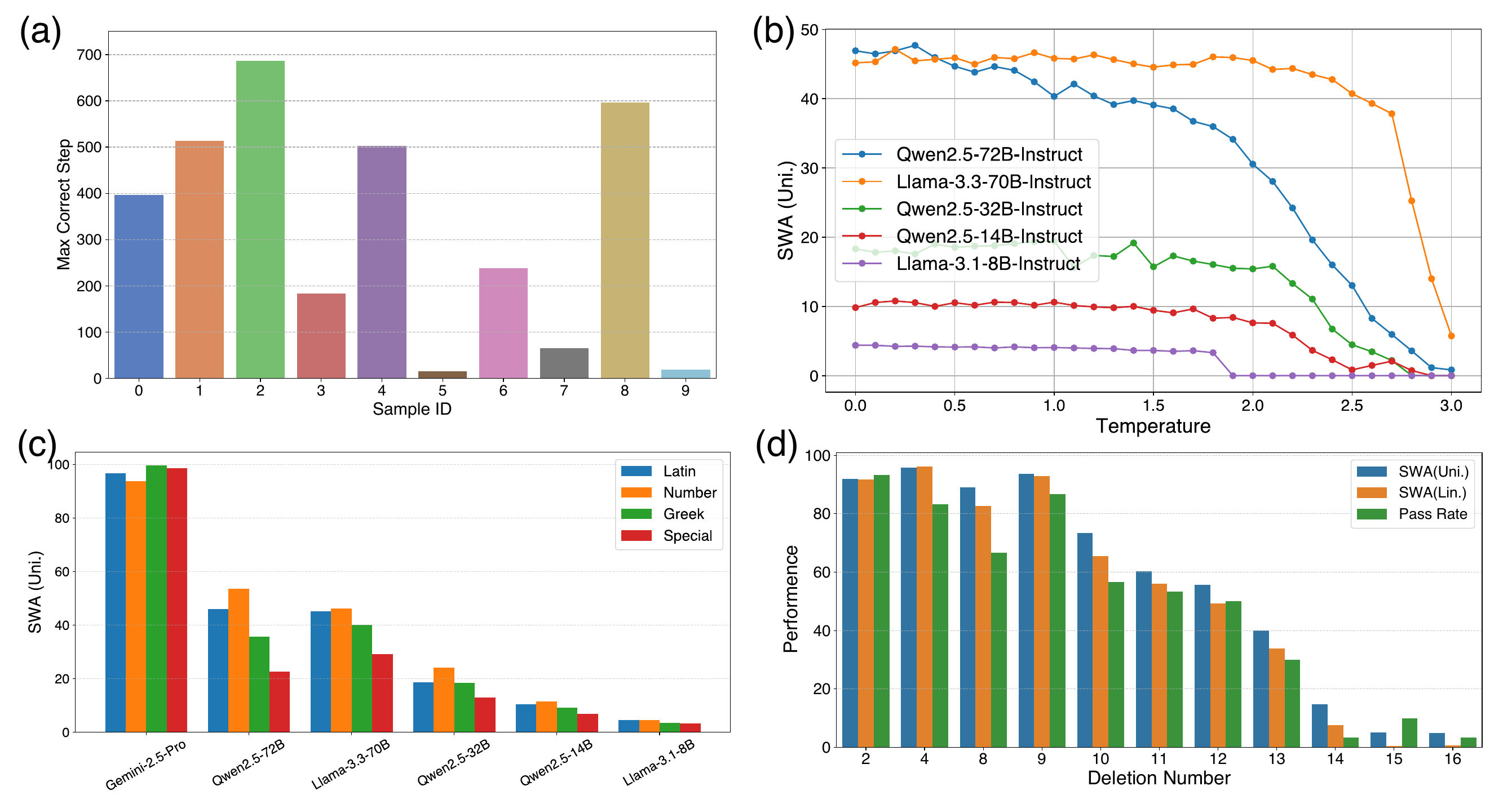}
    \vspace{-1em}
    \caption{Illustration of ablation results. (a) Maximum correct step achieved by Gemini under unbounded-step execution. (b) Impact of decoding temperature on performance. (c) Effect of alphabet types (Roman, Number, Greek, and Special) on model performance. (d) Task difficulty ablation by varying the number of deletions.}
    \label{fig:ablation}
\end{figure}

\paragraph{Unbounded-Step Ablation.} Gemini-2.5-pro demonstrates strong instruction-following capabilities, achieving a 94\% pass rate under a 30-step constraint. To further analyze the upper bound of its instruction-following ability, we conducted an unbounded-step evaluation. Using rejection sampling, we selected 10 samples and ran each for up to 1000 steps. Through budget forcing~\cite{muennighoff2025s1}, Gemini was prompted to continue generating until failure with maximum token limit of 1,048,576 tokens. The earliest failure occurred at step 16, and the latest at step 683, as shown in~\cref{fig:ablation}a. As an autoregressive model, Gemini inevitably fails with increasing steps due to its statistical nature, underscoring Gemini's computational limits.
\paragraph{Temperature ablation.} To examine the effect of temperature on model performance, we conduct an ablation study with temperature values ranging from 0 to 3.0, while setting the top-p value to 0.8, as shown in~\cref{fig:ablation}b. Performance remains stable between 0 and 1, but degrades at higher temperatures.
\paragraph{Alphabet ablation.}
To evaluate the robustness of large language models (LLMs) across diverse alphabet types, we conducted evaluations using four distinct character sets: \textbf{Roman letters} (e.g., \textit{a, b, c, d, e}), \textbf{Numerals} (e.g., \textit{1, 2, 3, 4, 5}), \textbf{Greek letters} (e.g.,~$\alpha$, $\beta$, $\gamma$, $\delta$, $\epsilon$), and \textbf{Special characters} (e.g.,~@, \#, \$, \%, \&). As illustrated in Figure~\cref{fig:ablation}c, while most models exhibit performance fluctuations depending on the character set, Gemini-2.5-Pro consistently achieves high accuracy across all categories, including the \textit{Special characters}, which typically present greater challenges. This robustness indicates that Gemini-2.5-Pro relies on underlying reasoning mechanisms rather than superficial statistical correlations.
% To evaluate the robustness of LLMs across different types of alphabets, we tested them using: Vanilla (e.g.,~a, b, c, d, e), Number (e.g.,~1, 2, 3, 4, 5), Greek (e.g.,~$\alpha$, $\beta$, $\gamma$, $\delta$, $\epsilon$), and Special characters (e.g.,~@, \#, \$, \%, \&). As shown in~\ref{fig:alphabet}, unlike the other models which exhibit more variability, Gemini-2.5-Pro demonstrates consistently high performance across all alphabet types, including the 'Special' characters that pose a challenge to others. This performance suggests underlying reasoning capabilities, rather than reliance solely on statistical correlations learned during training.
% 为了评估模型对于不同alphabet的鲁棒性，vanilla，number，greek，speical，vanilla是a,b,c,d,e, number是1,2,3,4,5 , greek 是αβγδε，speical是@#$%&，
% 如图3所示。
% Unlike the other models which show more variability, Gemini-2.5-Pro demonstrates consistently high performance across all alphabet types, including the 'Special' characters that challenge others. 说明了其推理的能力，而不只是基于统计。

\paragraph{Difficulty Ablation.} For TMBench, Gemini-2.5-pro achieves a 94\% pass rate within 30 steps, with successful simulations extending up to 686 steps, demonstrating its strong computational reasoning capabilities. To further assess this ability, we perform a difficulty ablation study with varying values of \(m\), as shown in~\cref{fig:ablation}d. The model maintains stable performance when the deletion number \(m\) is between 2 and 9, but performance declines steadily beyond \(m = 10\), approaching zero after \(m = 15\). This demonstrates that varying \(m\) produces a smooth difficulty gradient, underscoring TMBench’s scalability and effectiveness as a benchmarking tool.

% over the alphabet \(\Sigma = \{ E, 1, \#, \$, * \}\)
% Gemini-2.5-pro achieves SWA(Uni) 55.6, SWA(Lin) 49.3, and Pass rate 50\%.
% Results indicate that Gemini struggles with computational reasoning on this benchmark, reflecting limited ability to address similarly complex real-world problems. 
% These findings further indicate that statistical autoregressive models struggle to solve even relatively simple systems characterized by well-defined rules and structures.

% \subsection{SFT finetune}
% #######  m=2 m=3 
% vanilla
% SFT

% TMBench
% Qwen2.5-14B-Instruct_0.0: uni_acc=10.1, lin_acc=2.6 reasoning_pass=0
% sft_full_qwen2.5-14b_0.0: uni_acc=51.9, lin_acc=36.4 reasoning_pass=12
%

%greek sft_full_qwen2.5-14b_0.0: uni_acc=48.2, lin_acc=33.1 reasoning_pass=13
%number sft_full_qwen2.5-14b_0.0: uni_acc=43.8, lin_acc=27.3 reasoning_pass=9
%speicial sft_full_qwen2.5-14b_0.0: uni_acc=26.7, lin_acc=13.0 reasoning_pass=7

% \input{5_limitations}
\section{Conclusion}
In this paper, we propose computational reasoning as a basic ability of LLMs, the ability to strictly follow rules and accurately manage internal states for multi-step, a proxy capability that generalizes across tasks and is independent of domain-specific knowledge.
Motivated by this perspective, we introduce Turing Machine Bench (TMBench), a novel benchmark that leverages m-Tag System Simulation to evaluate the computational reasoning of LLMs. 
TMBench is characterized by four core attributes: (1) self-contained, knowledge-independent reasoning; (2) an interpretable, minimal multi-step structure; (3) adjustable task difficulty; and (4) computational generality grounded in Turing completeness, making it a rigorous proxy for basic ability.
In our experiments, we evaluate state-of-the-art open-source LLMs (ranging from 0.6B to 671B parameters), along with leading proprietary models such as Gemini and Grok on TMBench. The empirical results demonstrate that TMBench effectively captures computational reasoning capabilities across models, revealing a clear multi-step performance curve.
% Our results show that the leading model, Gemini-2.5-Pro, achieves a 94\% simulation pass rate, demonstrating strong general-purpose computational ability.
However, TMBench’s knowledge-agnostic nature does not hinder its correlation with real-world tasks. Further analysis of its performance against established reasoning benchmarks reveals a strong alignment with practical tasks, effectively bridging the gap between real-task evaluation and basic ability assessment. Notably, the correlation with reasoning benchmarks such as AIME and GPQA is higher than with knowledge-centric tasks like MMLU. 
To further investigate the benchmark's robustness and characteristics, we conduct several ablation studies, including unbounded step ablation, temperature ablation, alphabet ablation, and difficulty ablation. As an autoregressive model, Gemini inevitably exhibits performance degradation with increasing step counts due to its statistical nature, highlighting its computational limitations. However, Gemini maintains stable performance on variations involving numerals, Greek letters, and special characters, suggesting that its success relies on underlying reasoning mechanisms rather than superficial statistical correlations. By varying the number of deletions, TMBench provides a continuous and scalability spectrum of task difficulty.

\newpage
\bibliographystyle{unsrt}
\bibliography{meta/ref}

\begin{thebibliography}{10}

\bibitem{devlin2019bert}
Jacob Devlin, Ming-Wei Chang, Kenton Lee, and Kristina Toutanova.
\newblock Bert: Pre-training of deep bidirectional transformers for language understanding.
\newblock In {\em Proceedings of the 2019 conference of the North American chapter of the association for computational linguistics: human language technologies, volume 1 (long and short papers)}, pages 4171--4186, 2019.

\bibitem{rein2024gpqa}
David Rein, Betty~Li Hou, Asa~Cooper Stickland, Jackson Petty, Richard~Yuanzhe Pang, Julien Dirani, Julian Michael, and Samuel~R Bowman.
\newblock Gpqa: A graduate-level google-proof q\&a benchmark.
\newblock In {\em First Conference on Language Modeling}, 2024.

\bibitem{huang2024olympicarena}
Zhen Huang, Zengzhi Wang, Shijie Xia, Xuefeng Li, Haoyang Zou, Ruijie Xu, Run-Ze Fan, Lyumanshan Ye, Ethan Chern, Yixin Ye, et~al.
\newblock Olympicarena: Benchmarking multi-discipline cognitive reasoning for superintelligent ai.
\newblock {\em arXiv preprint arXiv:2406.12753}, 2024.

\bibitem{jimenez2023swe}
Carlos~E Jimenez, John Yang, Alexander Wettig, Shunyu Yao, Kexin Pei, Ofir Press, and Karthik Narasimhan.
\newblock Swe-bench: Can language models resolve real-world github issues?
\newblock {\em arXiv preprint arXiv:2310.06770}, 2023.

\bibitem{chen2021codex}
Mark Chen, Jerry Tworek, Heewoo Jun, Qiming Yuan, Henrique~Ponde de~Oliveira~Pinto, Jared Kaplan, Harri Edwards, Yuri Burda, Nicholas Joseph, Greg Brockman, Alex Ray, Raul Puri, Gretchen Krueger, Michael Petrov, Heidy Khlaaf, Girish Sastry, Pamela Mishkin, Brooke Chan, Scott Gray, Nick Ryder, Mikhail Pavlov, Alethea Power, Lukasz Kaiser, Mohammad Bavarian, Clemens Winter, Philippe Tillet, Felipe~Petroski Such, Dave Cummings, Matthias Plappert, Fotios Chantzis, Elizabeth Barnes, Ariel Herbert-Voss, William~Hebgen Guss, Alex Nichol, Alex Paino, Nikolas Tezak, Jie Tang, Igor Babuschkin, Suchir Balaji, Shantanu Jain, William Saunders, Christopher Hesse, Andrew~N. Carr, Jan Leike, Josh Achiam, Vedant Misra, Evan Morikawa, Alec Radford, Matthew Knight, Miles Brundage, Mira Murati, Katie Mayer, Peter Welinder, Bob McGrew, Dario Amodei, Sam McCandlish, Ilya Sutskever, and Wojciech Zaremba.
\newblock Evaluating large language models trained on code.
\newblock 2021.

\bibitem{la2024code}
Emanuele La~Malfa, Christoph Weinhuber, Orazio Torre, Fangru Lin, Samuele Marro, Anthony Cohn, Nigel Shadbolt, and Michael Wooldridge.
\newblock Code simulation challenges for large language models.
\newblock {\em arXiv preprint arXiv:2401.09074}, 2024.

\bibitem{wu2023next}
Shengqiong Wu, Hao Fei, Leigang Qu, Wei Ji, and Tat-Seng Chua.
\newblock Next-gpt: Any-to-any multimodal llm.
\newblock {\em arXiv preprint arXiv:2309.05519}, 2023.

\bibitem{boiko2023autonomous}
Daniil~A Boiko, Robert MacKnight, Ben Kline, and Gabe Gomes.
\newblock Autonomous chemical research with large language models.
\newblock {\em Nature}, 624(7992):570--578, 2023.

\bibitem{hendrycks2020measuring}
Dan Hendrycks, Collin Burns, Steven Basart, Andy Zou, Mantas Mazeika, Dawn Song, and Jacob Steinhardt.
\newblock Measuring massive multitask language understanding.
\newblock In {\em International Conference on Learning Representations}, 2020.

\bibitem{joshi2017triviaqa}
Mandar Joshi, Eunsol Choi, Daniel~S Weld, and Luke Zettlemoyer.
\newblock Triviaqa: A large scale distantly supervised challenge dataset for reading comprehension.
\newblock {\em arXiv preprint arXiv:1705.03551}, 2017.

\bibitem{arkoudas2023gpt}
Konstantine Arkoudas.
\newblock Gpt-4 can't reason.
\newblock {\em arXiv preprint arXiv:2308.03762}, 2023.

\bibitem{srivastava2023beyond}
BIG bench authors.
\newblock Beyond the imitation game: Quantifying and extrapolating the capabilities of language models.
\newblock {\em Transactions on Machine Learning Research}, 2023.

\bibitem{AIME2024}
MAA.
\newblock American invitational mathematics examination - aime.
\newblock In {\em American Invitational Mathematics Examination - AIME 2024}, February 2024.

\bibitem{zheng2021minif2f}
Kunhao Zheng, Jesse~Michael Han, and Stanislas Polu.
\newblock Minif2f: a cross-system benchmark for formal olympiad-level mathematics.
\newblock {\em arXiv preprint arXiv:2109.00110}, 2021.

\bibitem{azerbayev2023proofnet}
Zhangir Azerbayev, Bartosz Piotrowski, Hailey Schoelkopf, Edward~W Ayers, Dragomir Radev, and Jeremy Avigad.
\newblock Proofnet: Autoformalizing and formally proving undergraduate-level mathematics.
\newblock {\em arXiv preprint arXiv:2302.12433}, 2023.

\bibitem{lai2024executing}
Junyu Lai, Jiahe Xu, Yao Yang, Yunpeng Huang, Chun Cao, and Jingwei Xu.
\newblock Executing arithmetic: Fine-tuning large language models as turing machines.
\newblock {\em arXiv preprint arXiv:2410.07896}, 2024.

\bibitem{zellers2019hellaswag}
Rowan Zellers, Ari Holtzman, Yonatan Bisk, Ali Farhadi, and Yejin Choi.
\newblock Hellaswag: Can a machine really finish your sentence?
\newblock In {\em Proceedings of the 57th Annual Meeting of the Association for Computational Linguistics}, pages 4791--4800, 2019.

\bibitem{cobbe2021gsm}
Karl Cobbe, Vineet Kosaraju, Mohammad Bavarian, Mark Chen, Heewoo Jun, Lukasz Kaiser, Matthias Plappert, Jerry Tworek, Jacob Hilton, Reiichiro Nakano, et~al.
\newblock Training verifiers to solve math word problems.
\newblock {\em arXiv preprint arXiv:2110.14168}, 2021.

\bibitem{liang2022holistic}
Percy Liang, Rishi Bommasani, Tony Lee, Dimitris Tsipras, Dilara Soylu, Michihiro Yasunaga, Yian Zhang, Deepak Narayanan, Yuhuai Wu, Ananya Kumar, et~al.
\newblock Holistic evaluation of language models.
\newblock {\em arXiv preprint arXiv:2211.09110}, 2022.

\bibitem{zhong2023agieval}
Wanjun Zhong, Ruixiang Cui, Yiduo Guo, Yaobo Liang, Shuai Lu, Yanlin Wang, Amin Saied, Weizhu Chen, and Nan Duan.
\newblock Agieval: A human-centric benchmark for evaluating foundation models.
\newblock {\em arXiv preprint arXiv:2304.06364}, 2023.

\bibitem{chen2021evaluating}
Mark Chen, Jerry Tworek, Heewoo Jun, Qiming Yuan, Henrique Ponde de~Oliveira Pinto, Jared Kaplan, Harri Edwards, Yuri Burda, Nicholas Joseph, Greg Brockman, et~al.
\newblock Evaluating large language models trained on code.
\newblock {\em arXiv preprint arXiv:2107.03374}, 2021.

\bibitem{toxicchat}
Zi~Lin, Zihan Wang, Yongqi Tong, Yangkun Wang, Yuxin Guo, Yujia Wang, and Jingbo Shang.
\newblock {T}oxic{C}hat: Unveiling hidden challenges of toxicity detection in real-world user-{AI} conversation.
\newblock In Houda Bouamor, Juan Pino, and Kalika Bali, editors, {\em Findings of the Association for Computational Linguistics: EMNLP 2023}, pages 4694--4702, Singapore, December 2023. Association for Computational Linguistics.

\bibitem{chiang-lee-2023-large}
Cheng-Han Chiang and Hung-yi Lee.
\newblock Can large language models be an alternative to human evaluations?
\newblock In Anna Rogers, Jordan Boyd-Graber, and Naoaki Okazaki, editors, {\em Proceedings of the 61st Annual Meeting of the Association for Computational Linguistics (Volume 1: Long Papers)}, pages 15607--15631, Toronto, Canada, July 2023. Association for Computational Linguistics.

\bibitem{zheng2023judging}
Lianmin Zheng, Wei-Lin Chiang, Ying Sheng, Siyuan Zhuang, Zhanghao Wu, Yonghao Zhuang, Zi~Lin, Zhuohan Li, Dacheng Li, Eric Xing, Hao Zhang, Joseph~E. Gonzalez, and Ion Stoica.
\newblock Judging {LLM}-as-a-judge with {MT}-bench and chatbot arena.
\newblock In {\em Thirty-seventh Conference on Neural Information Processing Systems Datasets and Benchmarks Track}, 2023.

\bibitem{alpaca_eval}
Xuechen Li, Tianyi Zhang, Yann Dubois, Rohan Taori, Ishaan Gulrajani, Carlos Guestrin, Percy Liang, and Tatsunori~B. Hashimoto.
\newblock Alpacaeval: An automatic evaluator of instruction-following models.
\newblock \url{https://github.com/tatsu-lab/alpaca_eval}, 2023.

\bibitem{li2022competition}
Yujia Li, David Choi, Junyoung Chung, Nate Kushman, Julian Schrittwieser, R{\'e}mi Leblond, Tom Eccles, James Keeling, Felix Gimeno, Agustin Dal~Lago, et~al.
\newblock Competition-level code generation with alphacode.
\newblock {\em Science}, 378(6624):1092--1097, 2022.

\bibitem{huang2023competition}
Yiming Huang, Zhenghao Lin, Xiao Liu, Yeyun Gong, Shuai Lu, Fangyu Lei, Yaobo Liang, Yelong Shen, Chen Lin, Nan Duan, et~al.
\newblock Competition-level problems are effective llm evaluators.
\newblock {\em arXiv preprint arXiv:2312.02143}, 2023.

\bibitem{bai2022training}
Yuntao Bai, Andy Jones, Kamal Ndousse, Amanda Askell, Anna Chen, Nova DasSarma, Dawn Drain, Stanislav Fort, Deep Ganguli, Tom Henighan, et~al.
\newblock Training a helpful and harmless assistant with reinforcement learning from human feedback.
\newblock {\em arXiv preprint arXiv:2204.05862}, 2022.

\bibitem{ouyang2022training}
Long Ouyang, Jeff Wu, Xu~Jiang, Diogo Almeida, Carroll~L. Wainwright, Pamela Mishkin, Chong Zhang, Sandhini Agarwal, Katarina Slama, Alex Ray, John Schulman, Jacob Hilton, Fraser Kelton, Luke Miller, Maddie Simens, Amanda Askell, Peter Welinder, Paul Christiano, Jan Leike, and Ryan Lowe.
\newblock Training language models to follow instructions with human feedback, 2022.

\bibitem{touvron2023llama2}
Hugo Touvron, Louis Martin, Kevin Stone, Peter Albert, Amjad Almahairi, Yasmine Babaei, Nikolay Bashlykov, Soumya Batra, Prajjwal Bhargava, Shruti Bhosale, et~al.
\newblock Llama 2: Open foundation and fine-tuned chat models.
\newblock {\em arXiv preprint arXiv:2307.09288}, 2023.

\bibitem{lewkowycz2022solving}
Aitor Lewkowycz, Anders Andreassen, David Dohan, Ethan Dyer, Henryk Michalewski, Vinay Ramasesh, Ambrose Slone, Cem Anil, Imanol Schlag, Theo Gutman-Solo, et~al.
\newblock Solving quantitative reasoning problems with language models.
\newblock {\em Advances in Neural Information Processing Systems}, 35:3843--3857, 2022.

\bibitem{wei2022chain}
Jason Wei, Xuezhi Wang, Dale Schuurmans, Maarten Bosma, Fei Xia, Ed~Chi, Quoc~V Le, Denny Zhou, et~al.
\newblock Chain-of-thought prompting elicits reasoning in large language models.
\newblock {\em Advances in neural information processing systems}, 35:24824--24837, 2022.

\bibitem{nye2021show}
Maxwell Nye, Anders~Johan Andreassen, Guy Gur-Ari, Henryk Michalewski, Jacob Austin, David Bieber, David Dohan, Aitor Lewkowycz, Maarten Bosma, David Luan, et~al.
\newblock Show your work: Scratchpads for intermediate computation with language models.
\newblock {\em arXiv preprint arXiv:2112.00114}, 2021.

\bibitem{wang2022self}
Xuezhi Wang, Jason Wei, Dale Schuurmans, Quoc Le, Ed~Chi, Sharan Narang, Aakanksha Chowdhery, and Denny Zhou.
\newblock Self-consistency improves chain of thought reasoning in language models.
\newblock {\em arXiv preprint arXiv:2203.11171}, 2022.

\bibitem{lourie2021unicorn}
Nicholas Lourie, Ronan Le~Bras, Chandra Bhagavatula, and Yejin Choi.
\newblock Unicorn on rainbow: A universal commonsense reasoning model on a new multitask benchmark.
\newblock In {\em Proceedings of the AAAI Conference on Artificial Intelligence}, volume~35, pages 13480--13488, 2021.

\bibitem{geva2021did}
Mor Geva, Daniel Khashabi, Elad Segal, Tushar Khot, Dan Roth, and Jonathan Berant.
\newblock Did aristotle use a laptop? a question answering benchmark with implicit reasoning strategies.
\newblock {\em Transactions of the Association for Computational Linguistics}, 9:346--361, 2021.

\bibitem{cobbe2021training}
Karl Cobbe, Vineet Kosaraju, Mohammad Bavarian, Mark Chen, Heewoo Jun, Lukasz Kaiser, Matthias Plappert, Jerry Tworek, Jacob Hilton, Reiichiro Nakano, et~al.
\newblock Training verifiers to solve math word problems.
\newblock {\em arXiv preprint arXiv:2110.14168}, 2021.

\bibitem{hendrycks2021measuring}
Dan Hendrycks, Collin Burns, Saurav Kadavath, Akul Arora, Steven Basart, Eric Tang, Dawn Song, and Jacob Steinhardt.
\newblock Measuring mathematical problem solving with the math dataset.
\newblock {\em arXiv preprint arXiv:2103.03874}, 2021.

\bibitem{luo2023towards}
Man Luo, Shrinidhi Kumbhar, Mihir Parmar, Neeraj Varshney, Pratyay Banerjee, Somak Aditya, Chitta Baral, et~al.
\newblock Towards logiglue: A brief survey and a benchmark for analyzing logical reasoning capabilities of language models.
\newblock {\em arXiv preprint arXiv:2310.00836}, 2023.

\bibitem{han2022folio}
Simeng Han, Hailey Schoelkopf, Yilun Zhao, Zhenting Qi, Martin Riddell, Wenfei Zhou, James Coady, David Peng, Yujie Qiao, Luke Benson, et~al.
\newblock Folio: Natural language reasoning with first-order logic.
\newblock {\em arXiv preprint arXiv:2209.00840}, 2022.

\bibitem{patel2024multi}
Nisarg Patel, Mohith Kulkarni, Mihir Parmar, Aashna Budhiraja, Mutsumi Nakamura, Neeraj Varshney, and Chitta Baral.
\newblock Multi-logieval: Towards evaluating multi-step logical reasoning ability of large language models.
\newblock {\em arXiv preprint arXiv:2406.17169}, 2024.

\bibitem{parmar2024towards}
Mihir Parmar, Nisarg Patel, Neeraj Varshney, Mutsumi Nakamura, Man Luo, Santosh Mashetty, Arindam Mitra, and Chitta Baral.
\newblock Towards systematic evaluation of logical reasoning ability of large language models.
\newblock {\em arXiv preprint arXiv:2404.15522}, 2024.

\bibitem{liu2020logiqa}
Jian Liu, Leyang Cui, Hanmeng Liu, Dandan Huang, Yile Wang, and Yue Zhang.
\newblock Logiqa: A challenge dataset for machine reading comprehension with logical reasoning.
\newblock {\em arXiv preprint arXiv:2007.08124}, 2020.

\bibitem{tafjord2020proofwriter}
Oyvind Tafjord, Bhavana~Dalvi Mishra, and Peter Clark.
\newblock Proofwriter: Generating implications, proofs, and abductive statements over natural language.
\newblock {\em arXiv preprint arXiv:2012.13048}, 2020.

\bibitem{zellers2019recognition}
Rowan Zellers, Yonatan Bisk, Ali Farhadi, and Yejin Choi.
\newblock From recognition to cognition: Visual commonsense reasoning.
\newblock In {\em Proceedings of the IEEE/CVF conference on computer vision and pattern recognition}, pages 6720--6731, 2019.

\bibitem{xiao2021next}
Junbin Xiao, Xindi Shang, Angela Yao, and Tat-Seng Chua.
\newblock Next-qa: Next phase of question-answering to explaining temporal actions.
\newblock In {\em Proceedings of the IEEE/CVF conference on computer vision and pattern recognition}, pages 9777--9786, 2021.

\bibitem{lu2022learn}
Pan Lu, Swaroop Mishra, Tanglin Xia, Liang Qiu, Kai-Wei Chang, Song-Chun Zhu, Oyvind Tafjord, Peter Clark, and Ashwin Kalyan.
\newblock Learn to explain: Multimodal reasoning via thought chains for science question answering.
\newblock {\em Advances in Neural Information Processing Systems}, 35:2507--2521, 2022.

\bibitem{handa2024actionreasoningbench}
Divij Handa, Pavel Dolin, Shrinidhi Kumbhar, Tran~Cao Son, and Chitta Baral.
\newblock Actionreasoningbench: Reasoning about actions with and without ramification constraints.
\newblock {\em arXiv preprint arXiv:2406.04046}, 2024.

\bibitem{suppes2012axiomatic}
Patrick Suppes.
\newblock {\em Axiomatic set theory}.
\newblock Courier Corporation, 2012.

\bibitem{russell2020principles}
Bertrand Russell.
\newblock {\em Principles of mathematics}.
\newblock Routledge, 2020.

\bibitem{feynman1963feynman}
Richard~Phillips Feynman.
\newblock The feynman lectures on physics.
\newblock {\em (No Title)}, 1:46, 1963.

\bibitem{conway1970game}
John Conway et~al.
\newblock The game of life.
\newblock {\em Scientific American}, 223(4):4, 1970.

\bibitem{moura2021lean}
Leonardo~de Moura and Sebastian Ullrich.
\newblock The lean 4 theorem prover and programming language.
\newblock In {\em Automated Deduction--CADE 28: 28th International Conference on Automated Deduction, Virtual Event, July 12--15, 2021, Proceedings 28}, pages 625--635. Springer, 2021.

\bibitem{turing1936computable}
Alan~Mathison Turing et~al.
\newblock On computable numbers, with an application to the entscheidungsproblem.
\newblock {\em J. of Math}, 58(345-363):5, 1936.

\bibitem{Wang1971-WANTSA-3}
Hao Wang, John Cocke, Marvin Minsky, and Stephen~A. Cook.
\newblock Tag systems and lag systems.
\newblock {\em Journal of Symbolic Logic}, 36(2):344--344, 1971.

\bibitem{UTM-2tag}
John Cocke and Marvin Minsky.
\newblock Universality of tag systems with p = 2.
\newblock {\em J. ACM}, 11(1):15–20, January 1964.

\bibitem{post1943formal}
Emil~L Post.
\newblock Formal reductions of the general combinatorial decision problem.
\newblock {\em American journal of mathematics}, 65(2):197--215, 1943.

\bibitem{lightman2023lets}
Hunter Lightman, Vineet Kosaraju, Yura Burda, Harri Edwards, Bowen Baker, Teddy Lee, Jan Leike, John Schulman, Ilya Sutskever, and Karl Cobbe.
\newblock Let's verify step by step.
\newblock {\em arXiv preprint arXiv:2305.20050}, 2023.

\bibitem{wang2024mmlu}
Yubo Wang, Xueguang Ma, Ge~Zhang, Yuansheng Ni, Abhranil Chandra, Shiguang Guo, Weiming Ren, Aaran Arulraj, Xuan He, Ziyan Jiang, et~al.
\newblock Mmlu-pro: A more robust and challenging multi-task language understanding benchmark.
\newblock In {\em The Thirty-eight Conference on Neural Information Processing Systems Datasets and Benchmarks Track}, 2024.

\bibitem{dubey2024llama}
Abhimanyu Dubey, Abhinav Jauhri, Abhinav Pandey, Abhishek Kadian, Ahmad Al-Dahle, Aiesha Letman, Akhil Mathur, Alan Schelten, Amy Yang, Angela Fan, et~al.
\newblock The llama 3 herd of models.
\newblock {\em arXiv preprint arXiv:2407.21783}, 2024.

\bibitem{yang2024qwen2}
An~Yang, Baosong Yang, Beichen Zhang, Binyuan Hui, Bo~Zheng, Bowen Yu, Chengyuan Li, Dayiheng Liu, Fei Huang, Haoran Wei, et~al.
\newblock Qwen2. 5 technical report.
\newblock {\em arXiv preprint arXiv:2412.15115}, 2024.

\bibitem{wang2024qwen2}
Peng Wang, Shuai Bai, Sinan Tan, Shijie Wang, Zhihao Fan, Jinze Bai, Keqin Chen, Xuejing Liu, Jialin Wang, Wenbin Ge, et~al.
\newblock Qwen2-vl: Enhancing vision-language model's perception of the world at any resolution.
\newblock {\em arXiv preprint arXiv:2409.12191}, 2024.

\bibitem{qvq-72b-preview}
Qwen Team.
\newblock Qvq: To see the world with wisdom, December 2024.

\bibitem{team2025gemma}
Gemma Team, Aishwarya Kamath, Johan Ferret, Shreya Pathak, Nino Vieillard, Ramona Merhej, Sarah Perrin, Tatiana Matejovicova, Alexandre Ram{\'e}, Morgane Rivi{\`e}re, et~al.
\newblock Gemma 3 technical report.
\newblock {\em arXiv preprint arXiv:2503.19786}, 2025.

\bibitem{team2025sky}
NovaSky Team.
\newblock Sky-t1: Train your own o1 preview model within \$450.
\newblock 2025.

\bibitem{gemini2025pro}
Google DeepMind.
\newblock Gemini 2.5 pro, 2025.
\newblock Accessed: 2025-03-28.

\bibitem{grok3}
{xAI}.
\newblock Grok-3, 2025.
\newblock Accessed: 2025-02-19.

\bibitem{claude37sonnet}
{Anthropic}.
\newblock Claude 3.7 sonnet thinking, 2025.
\newblock Accessed: 2025-02-25.

\bibitem{doubao15pro}
{Doubao}.
\newblock Doubao 1.5 pro, 2025.
\newblock Accessed: 2025-01-22.

\bibitem{openai-o1-systemcard}
OpenAI.
\newblock Openai o1 system card, Sep. 2024.
\newblock (Accessed on 2024/09/23).

\bibitem{wolf2020transformers}
Thomas Wolf, Lysandre Debut, Victor Sanh, Julien Chaumond, Clement Delangue, Anthony Moi, Pierric Cistac, Tim Rault, R{\'e}mi Louf, Morgan Funtowicz, et~al.
\newblock Transformers: State-of-the-art natural language processing.
\newblock In {\em Proceedings of the 2020 conference on empirical methods in natural language processing: system demonstrations}, pages 38--45, 2020.

\bibitem{kwon2023efficient}
Woosuk Kwon, Zhuohan Li, Siyuan Zhuang, Ying Sheng, Lianmin Zheng, Cody~Hao Yu, Joseph~E. Gonzalez, Hao Zhang, and Ion Stoica.
\newblock Efficient memory management for large language model serving with pagedattention.
\newblock In {\em Proceedings of the ACM SIGOPS 29th Symposium on Operating Systems Principles}, 2023.

\bibitem{wei2022emergent}
Jason Wei, Yi~Tay, Rishi Bommasani, Colin Raffel, Barret Zoph, Sebastian Borgeaud, Dani Yogatama, Maarten Bosma, Denny Zhou, Donald Metzler, et~al.
\newblock Emergent abilities of large language models.
\newblock {\em arXiv preprint arXiv:2206.07682}, 2022.

\bibitem{muennighoff2025s1}
Niklas Muennighoff, Zitong Yang, Weijia Shi, Xiang~Lisa Li, Li~Fei-Fei, Hannaneh Hajishirzi, Luke Zettlemoyer, Percy Liang, Emmanuel Cand{\`e}s, and Tatsunori Hashimoto.
\newblock s1: Simple test-time scaling.
\newblock {\em arXiv preprint arXiv:2501.19393}, 2025.

\end{thebibliography}

%%%%%%%%%%%%%%%%%%%%%%%%%%%%%%%%%%%%%%%%%%%%%%%%%%%%%%%%%%%%
\newpage
\appendix
\section{Theory.}
\label{sec:prove}
% You can have as much text here as you want. The main body must be at most $8$ pages long.
% For the final version, one more page can be added.
% If you want, you can use an appendix like this one.  

% The $\mathtt{\backslash onecolumn}$ command above can be kept in place if you prefer a one-column appendix, or can be removed if you prefer a two-column appendix.  Apart from this possible change, the style (font size, spacing, margins, page numbering, etc.) should be kept the same as the main body.
In this section, we demonstrate that the m-tag (m>1) system constitutes a universal Turing machine, following the construction presented in~\cite{UTM-2tag}, with minor corrections.
\subsection{Representation of a Turing Machine}
We begin by introducing the definition and step operation for the Turing Machine.
Consider a Turing Machine with a binary alphabet \( \Gamma = \{0, 1\} \), where each state \( Q_i \) is defined by a quadruple \( (S_i, D_i, Q_{i0}, Q_{i1}) \), corresponding to the transitions for the symbols 0 and 1.
A single step of the Turing Machine, when in state \( Q_i \), is performed as follows:

\begin{enumerate}
    \item Write the symbol \( S_i \) onto the current tape cell.
    \item Move the tape head one cell in direction \( D_i \), where \( D_i \in \{L, R\} \).
    \item Read the symbol \( S' \) from the cell currently.
    \item Transition to the next state based on the symbol read \( S' \):
    \begin{itemize}
        \item If \( S' = 0 \), the next state is \( Q_{i0} \).
        \item If \( S' = 1 \), the next state is \( Q_{i1} \).
    \end{itemize}
\end{enumerate}

In this specific transition rule, the symbol written \( S_i \) and the tape head direction \( D_i \) are determined solely by the current state \( Q_i \). The subsequent state transition depends on the symbol \( S' \) read from the tape after the move: the machine transitions to state \( Q_{i0} \) if \( S' = 0 \), and to state \( Q_{i1} \) if \( S' = 1 \).

An instantaneous description of the machine with state \( Q_i \) at any given step is represented as:
\[
\cdots a_3a_2a_1a_0 \alpha b_0b_1b_2b_3 \cdots
\]

where \( Q_i \) is the state after reading \( \alpha \). The complete instantaneous description can be described by a \textit{triplet}:
\[
(Q,M,N) = (Q_i,\sum_{i=0}^{\infty}a_i2^i,\sum_{i=0}^{\infty}b_i2^i)
\]

where \( M \) is the string of symbols to the left of \( \alpha \), \( N \) is the string of symbols to the right of \( \alpha \). The symbol \( \alpha \) is the current symbol being read, which will soon be replaced by \( S_i \) depending on the transition rule.

For a rightward movement, the next state \( Q' \) depends on the value of \( S_i \). If \( S_i = 0 \), then \( Q' = Q_{i0} \), and if \( S_i = 1 \), then \( Q' = Q_{i1} \).

The updating of the strings follows the rules:
\[
M \leftarrow 2M + S_i
\]
\[
N \leftarrow \left\lfloor \frac{N}{2} \right\rfloor
\]

In the case of a rightward movement, only the values of $M$ and $N$ are swapped, due to the symmetry of the process. For the sake of simplicity, we assume here that the focus is on the rightward move, without loss of generality.

\subsection{Equivalent Construction with a Tag System}

Given the instantaneous description of the Turing Machine as a triplet \( (Q_i, M, N) \), we can construct a corresponding string using the following form:
\[
A_ix_i(\alpha_ix_i)^M B_ix_i(\beta_ix_i)^N
\]

For simplicity and clarity, we will omit the subscript \( i \) in the notation and rewrite it as:
\[
A x (\alpha x)^M B x (\beta x)^N
\]
% The goal is to evolve this string according to the production rules to produce the next configuration:
% \[
% A_0 x_0 (\alpha_0 x_0)^M B_0 x_0 (\beta_0 x_0)^{\frac{N}{2}}
% \]

\subsubsection*{Step 0 (Initial)}
\begin{equation}
A x (\alpha x)^M B x (\beta x)^N
\end{equation}
\subsubsection*{Step 1}
Rules:
\[
A \rightarrow 
\begin{cases}
Cx, & \text{if } S = 0 \\
Cxcx, & \text{if } S = 1
\end{cases},
\quad
\alpha \rightarrow cxcx
\]

% In this step, the transformation rules are applied based on the value of $S$:
% \begin{itemize}
%   \item If \( S = 0 \), \( A \rightarrow Cx \)
%   \item If \( S = 1 \), \( A \rightarrow Cxcx \)
%   \item \( \alpha \rightarrow cxcx \)
% \end{itemize}
Updated:
\begin{equation}
B x (\beta x)^N C x (c x)^{M'}
\end{equation}
where
\[
M' = 
\begin{cases}
2M, & \text{if } S = 0 \\
2M + 1, & \text{if } S = 1
\end{cases}
\]
% where \( M' = 2M \) or \( 2M + 1 \), depending on \( S = 0 \) or \( S = 1 \).

\subsubsection*{Step 2}
Rules:
\[
B \rightarrow S, \quad \beta \rightarrow s
\]
Updated:
\begin{equation}
C x (c x)^{M'} S (s)^N
\end{equation}

\subsubsection*{Step 3}
Rules:
\[
C \rightarrow D_1 D_0, \quad c \rightarrow d_1 d_0
\]
Updated:
\begin{equation}
S (s)^N D_1 D_0 (d_1 d_0)^{M'}
\end{equation}

\subsubsection*{Step 4}
Rules:
\[
S \rightarrow T_1 T_0, \quad s \rightarrow t_1 t_0
\]
Based on the parity of \( N \), the updated string takes different forms:

If \( N \) is odd:
  \[
  D_1 D_0 (d_1 d_0)^{M'} T_1 T_0 (t_1 t_0)^{\frac{N-1}{2}},
  \]
If \( N \) is even:
  \[
  D_0 (d_1 d_0)^{M'} T_1 T_0 (t_1 t_0)^{\frac{N}{2}}.
  \]

\subsubsection*{Step 5}

If \( N \) is odd:
\[
D_1 \rightarrow A_1 x_1, \quad d_1 \rightarrow a_1 x_1
\]
Then, the resulting string is:
\begin{equation}
T_1 T_0 (t_1 t_0)^{\frac{N-1}{2}} A_1 x_1 (\alpha_1 x_1)^{M'}
\end{equation}

If \( N \) is even:
\[
D_0 \rightarrow x_0 A_0 x_0, \quad d_0 \rightarrow a_0 x_0
\]
Then, the resulting string is:
\begin{equation}
T_0 (t_1 t_0)^{\frac{N}{2}} x_0 A_0 x_0 (\alpha_0 x_0)^{M'}
\end{equation}

\subsubsection*{Step 6 (Final)}

If \( N \) is odd:
\[
T_1 \rightarrow B_1 x_1, \quad t_1 \rightarrow \beta_1 x_1
\]
Then,
\begin{equation}
A_1 x_1 (\alpha_1 x_1)^{M'} B_1 x_1 (\beta_1 x_1)^{\frac{N-1}{2}}
\end{equation}

If \( N \) is even:
\[
T_0 \rightarrow B_0 x_0, \quad t_0 \rightarrow \beta_0 x_0
\]
Then,
\begin{equation}
A_0 x_0 (\alpha_0 x_0)^{M'} B_0 x_0 (\beta_0 x_0)^{\frac{N}{2}}
\end{equation}
This completes the proof that a single computation step of a Turing machine is successfully simulated by a sequence of transitions in the 2-tag system.

% \newpage
\section{Experiment Details}
% \subsection{Datasets details}
% \subsection{Implementation details}
% \input{tabs/num_queue}
% \input{tabs/num_rule}
\subsection{Dataset Details}
\textbf{AIME2024}~\cite{AIME2024} is a collection of problems from the 2024 American Invitational Mathematics Examination. It contains 30 high-school level mathematics problems requiring creative problem solving and deep mathematical insight. These problems are typically used to select top students for the USA Mathematical Olympiad. The dataset emphasizes algebraic manipulation, mathematical reasoning, and non-routine problem solving. \\
\textbf{Math500}~\cite{lightman2023lets} is a representative benchmark consisting of 500 mathematics problems, uniformly sampled at random from the MATH\cite{hendrycks2021measuring} evaluation dataset by OpenAI. It includes problems of varying difficulty across topics such as algebra, geometry, and number theory.\\
\textbf{GPQA Diamond}~\cite{rein2024gpqa} is a high-difficulty subset of Graduate-Level Google-Proof Q\&A Benchmark (GPQA), consisting of 198 multiple-choice questions in biology, physics, and chemistry. It includes only questions where both experts answer correctly and the majority of non-experts answer incorrectly to ensure highest quality.
\\
\textbf{MMLU Pro}~\cite{wang2024mmlu} is a professional-level extension of the original Massive Multitask Language Understanding (MMLU) benchmark~\cite{hendrycks2020measuring}, comprising expert-level questions across 57 domains, including mathematics, history, computer science, law, medicine, engineering, the natural sciences, and more. It enhances the original benchmark by introducing more challenging, reasoning-intensive questions, expanding answer choices from four to ten, and removing trivial or noisy items. \\
% \textbf{TMBench} is 

\subsection{Implementation Details}
\paragraph{Environment.}
Our method is implemented with Python 3.10.16, CUDA 12.4, PyTorch 2.6.0 and vLLM 0.8.5. The required libraries are specified in the \texttt{requirements.txt} file provided in the repository. The experiments are performed on a machine with 96 vCPUs (2.90 GHz) from Intel Xeon processors, eight NVIDIA A100-80GB GPUs, and 1024 GB of RAM.
\paragraph{Inference Configuration.}
For API-only models, we use their official APIs and set a maximum token length limit accordingly.
For open-source models that exceed the GPU memory limits of a single machine, such as LLaMA 4 Maverick, DeepSeek-V3, and DeepSeek-R1, we rely on APIs provided by Alibaba Cloud and DeepSeek.
For all other models, including Qwen, LLaMA, and Gemma, we use vLLM as the inference backend, which is highly efficient. We apply a sampling strategy based on greedy decoding with a maximum token limit of 16,384.

\paragraph{Prompt.} We adopt the following 1-shot reasoning prompt template for evaluation, where the shot example is used to guide the output format extraction.

\begin{promptbox}
<|im\_start|>user \\
Simulate a m-tag system. Your task is to simulate each transition step-by-step and provide the queue's state at each step. Follow the rules and examples closely, and stop upon reaching the halt condition or 30 steps. Do not generate additional examples or new problems. No code.

\#\# Rules for Simulation:\\
1. In each transition, the machine performs the following steps:\\
\hspace*{0.5em}	- If the queue length is less than m, halt\\
\hspace*{0.5em}	- Read the head symbol of queue\\
\hspace*{0.5em}	- Append symbols to the tail based on the head symbol and the corresponding transition rule\\
\hspace*{0.5em}	- Delete m symbols from the head of the queue\\

2. The machine halt if:\\
\hspace*{0.5em} - The queue's length is less than m.\\

\#\# Example: \\
m: 2\\
Alphabet: {{A, B, C}} \\
Init: [B C A]\\
Transition rules:\\
A : C A C\\
B : A\\
C : B\\
Simulation steps:\\
\#\#\# step 0:\\
   - Action: Init\\
   - Queue State: [B C A]\\

\#\#\# step 1:\\
   - Head Symbol: B\\
   - Action: Append A to the end of the queue. Remove B C from the head.\\
   - Queue State: [A A]\\
...\\
\#\#\# step 4:\\
   - Head Symbol: C\\
   - Action: Append B to the end of the queue. Remove C B from the head.\\
   - Queue State: [B] <halt>\\

---

\#\# The Only Problem to Solve:\\
m: \{m\}\\
Alphabet: \{ALPHABET\}\\
Init: \{INIT\}\\
Transition Rules:\\
\{RULES\}\\
Simulation steps:\\
<|im\_end|>

<|im\_start|>assistant
\end{promptbox}

\newpage
\paragraph{Case.} The response generated by Gemini-2.5-Pro, showcasing its impressive capabilities in instruction following and reasoning.

\begin{promptbox}
% \#\# The Only Problem to Solve:\\
% m: 2\\\\
% Alphabet: {A, B, C, D, E}\\
% Init: [D B E E A A]\\
% Transition Rules:\\
% A : C E\\
% B : B\\
% C : D C B B\\
% D : A E C D\\
% E : D\\
% Simulation steps:\\
<|im\_start|>assistant\\
Here is the step-by-step simulation:\\
m: 2\\
Alphabet: {A, B, C, D, E}\\
Init: [B D D]\\
Transition Rules:\\
A : C\\
B : E C E C\\
C : B B B A\\
D : D B B\\
E : A E E E\\
Simulation steps:\\
\#\#\# step 0:\\
- Action: Init\\
- Queue State: [B D D]\\
\#\#\# step 1:\\
- Head Symbol: B\\
- Action: Append E C E C to the end of the queue. Remove B D from the head.\\
- Queue State: [D E C E C]\\
\#\#\# step 2:\\
- Head Symbol: D\\
- Action: Append D B B to the end of the queue. Remove D E from the head.\\
- Queue State: [C E C D B B]\\
% \#\#\# step 3:\\
% - Head Symbol: C\\
% - Action: Append B B B A to the end of the queue. Remove C E from the head.\\
% - Queue State: [C D B B B B B A]\\[C E C D B B]
...\\
\#\#\# step 18:\\
- Head Symbol: E\\
- Action: Append A E E E to the end of the queue. Remove E C from the head.\\
- Queue State: [E C A E E E A E E E A E E E A E E E A E E E A E E E A E E E A E E E A E E E]\\
\#\#\# step 19:\\
- Head Symbol: E\\
- Action: Append A E E E to the end of the queue. Remove E C from the head.\\
- Queue State: [A E E E A E E E A E E E A E E E A E E E A E E E A E E E A E E E A E E E \textcolor{red}{\_ \_ \_ \_}] (Answer)\\
- Queue State: [A E E E A E E E A E E E A E E E A E E E A E E E A E E E A E E E A E E E \textcolor{deepgreen}{A E E E}] (Ground Truth)\\
...\\
\end{promptbox}

% \begin{quote}
% 

% \texttt{\#\#}  Rules for Simulation: \ldots

% \texttt{\#\#} The Only Problem to Solve: \\
% Alphabet: \texttt{\{ALPHABET\}} \\
% Deletion number: \texttt{\{delete\_num\}} \\
% Init: [{\texttt{\{INIT\}}}] \\
% Transition Rules: \\
% \texttt{\{RULES\}} \\
% Simulation steps: \\
% \texttt{<LLM>}
% \end{quote}

% You can have as much text here as you want. The main body must be at most $8$ pages long.
% For the final version, one more page can be added.
% If you want, you can use an appendix like this one.  

% The $\mathtt{\backslash onecolumn}$ command above can be kept in place if you prefer a one-column appendix, or can be removed if you prefer a two-column appendix.  Apart from this possible change, the style (font size, spacing, margins, page numbering, etc.) should be kept the same as the main body.

\clearpage
\subsection{More Results}
\begin{figure}[h]
  \centering
    \includegraphics[width=1.0\linewidth]{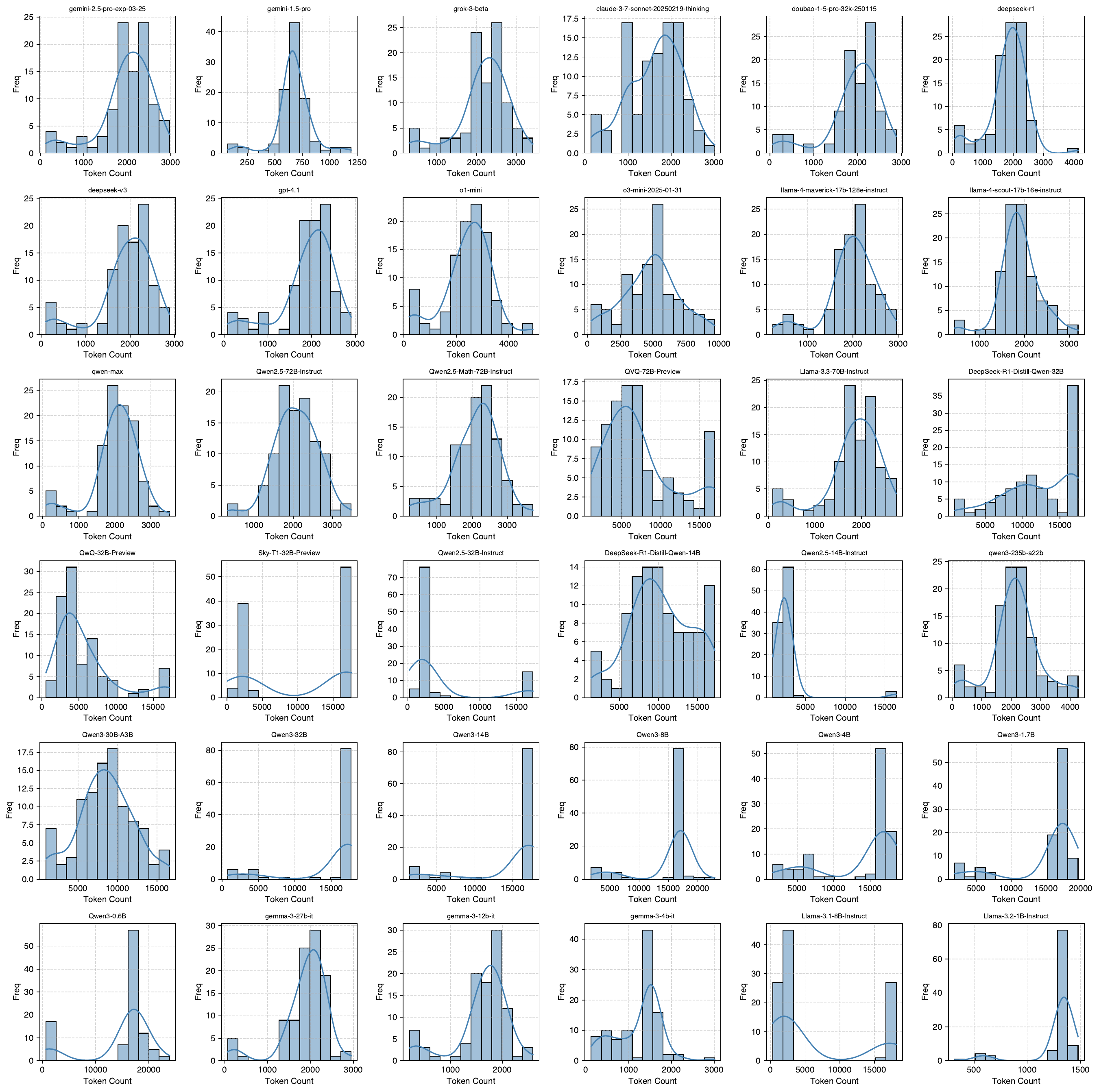}
    % \vspace{-1em}
    \caption{Illustration of token distrubition.}
    \label{fig:token_distribution}
\end{figure}

\paragraph{Token Distrubition.} We present the token distribution among 36 models on TMBench, as illustrated in~\cref{fig:token_distribution}. A higher token count is not necessarily better. Models that perform more reasoning tend to generate more tokens, but this does not always lead to improved performance. In some cases, the large number of tokens results from repetitive output, commonly referred to as the repetition problem. For state-of-the-art models such as Gemini 2.5 Pro, Grok, and Claude, their token distributions are similar.

\begin{figure}[h]
  \centering
    \includegraphics[width=1.0\linewidth]{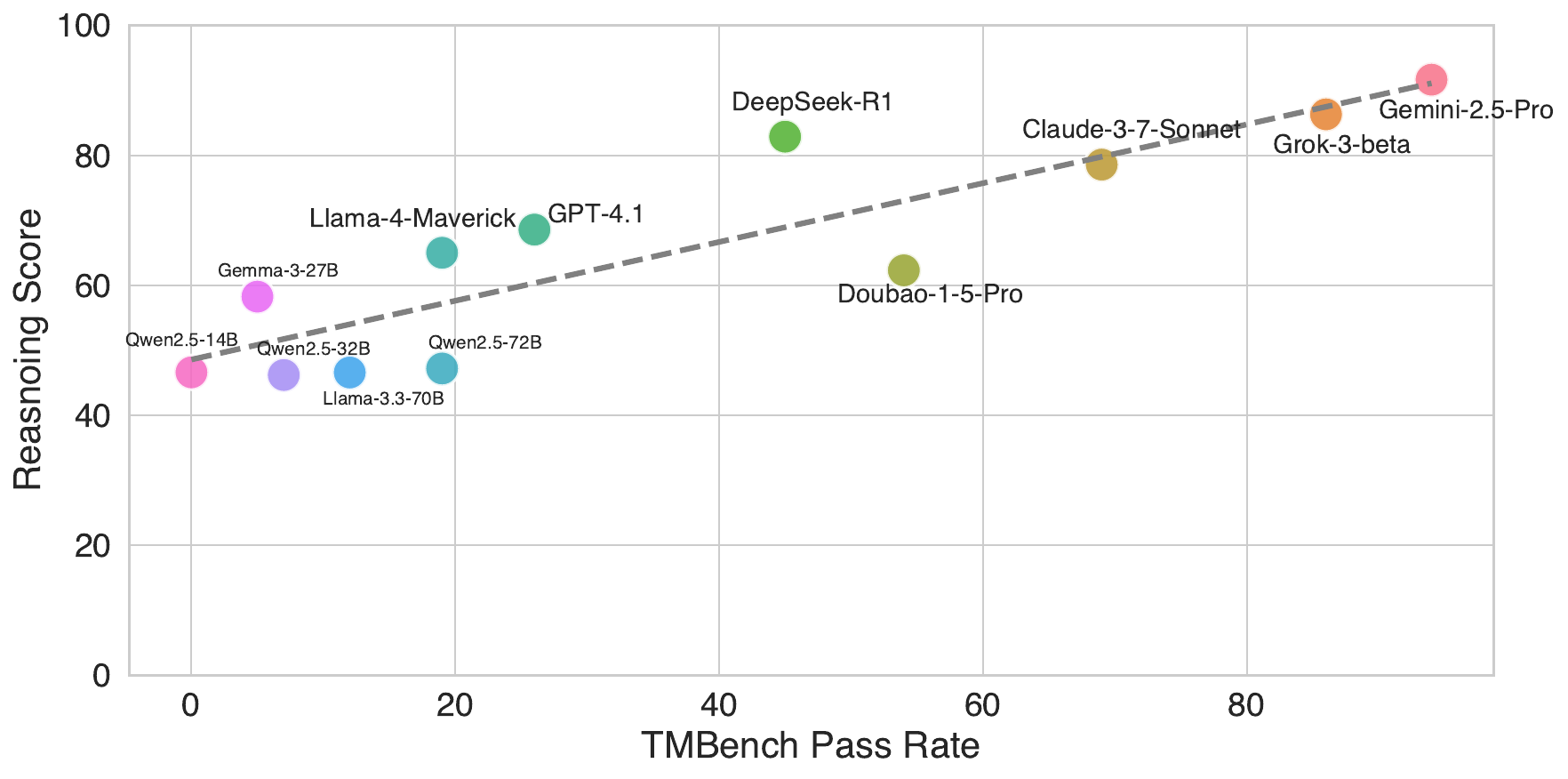}
    % \vspace{-1em}
    \caption{Correlation between TMBench Pass Rate (computational reasoning ability) and Reasoning Score (averaged across AIME2024, MATH500, and GPQA) among LLMs, with a Pearson correlation coefficient of 0.882 with p=1.49e-04, demonstrating the connection between TMBench as an abstract rule-based system simulation and real-world reasoning problem.}
    \label{fig:correlation}
\end{figure}

\paragraph{Benchmark Correlation.} We compute the average score across AIME2024, MATH500, and GPQA Diamond,
obtaining a Pearson correlation coefficient of 0.882 with the TMBench pass rate, as shown in~\cref{fig:correlation}. This strong correlation suggests that advanced reasoning performance is closely associated with computational reasoning ability.

% \clearpage
% \input{section/7_checklist}

\end{document}